\documentclass[11pt,a4paper]{article}
\usepackage[utf8]{inputenc}

\usepackage[american]{babel}

\usepackage[superscript,nomove]{cite}
\usepackage{amsmath,amssymb,amsfonts}
\usepackage{algorithmic}
\usepackage{algorithm}

\usepackage{graphicx}
\usepackage{textcomp}
\usepackage{booktabs}
\usepackage{numprint}
\usepackage{authblk}
\def\BibTeX{{\rm B\kern-.05em{\sc i\kern-.025em b}\kern-.08em
    T\kern-.1667em\lower.7ex\hbox{E}\kern-.125emX}}
\usepackage{listings}
\usepackage{booktabs}
\usepackage{multirow}
\usepackage{makecell}
\usepackage{placeins}
\usepackage{comment}
\usepackage{soul}
\usepackage[hyphens]{url}
\usepackage{breakurl}
\usepackage{float}
\usepackage[breaklinks]{hyperref}
\usepackage{subfig}
\usepackage{longtable}
\usepackage[table]{xcolor}
\usepackage{multirow}
\usepackage[font={small},labelfont={bf}]{caption}
\usepackage[misc]{ifsym}
\usepackage{sectsty}%
\usepackage{titlecaps}%
\usepackage[most]{tcolorbox}
\usepackage{booktabs}  %
\usepackage{array}  %
\usepackage[margin=1in]{geometry}
\definecolor{darkblue}{rgb}{0.0, 0.0, 0.6}
\definecolor{darkerblue}{rgb}{0.0, 0.0, 0.4}
\definecolor{darkcyan}{rgb}{0.0, 0.6, 0.6}
\definecolor{darkercyan}{rgb}{0.0, 0.4, 0.4}
\definecolor{darkerbluegray}{RGB}{67, 106, 161}
\definecolor{darkestbluegray}{RGB}{27, 66, 121}
\definecolor{darkred}{rgb}{0.6, 0.0, 0.0}
\definecolor{darkerred}{rgb}{0.4, 0.0, 0.0}
\definecolor{darkgreen}{rgb}{0.0, 0.4, 0.0}
\definecolor{darkergreen}{rgb}{0.0, 0.3, 0.0}
\definecolor{darksalmon}{rgb}{0.71, 0.39, 0.28}
\definecolor{darkersalmon}{rgb}{0.61, 0.29, 0.18}
\definecolor{darkpurple}{rgb}{0.5, 0.0, 0.5}
\definecolor{darkerpurple}{rgb}{0.4, 0.0, 0.4}
\definecolor{darkteal}{rgb}{0.0, 0.4, 0.4}
\definecolor{darkerteal}{rgb}{0.0, 0.3, 0.3}
\definecolor{darklavender}{rgb}{0.5294, 0.3020, 0.4745}
\definecolor{darkviolet}{RGB}{122, 65, 145}
\definecolor{darkyellow}{rgb}{0.5, 0.5, 0.0}
\definecolor{darkpink}{rgb}{0.7, 0.0, 0.5}
\definecolor{darkerpink}{rgb}{0.6, 0.0, 0.4}
\definecolor{black}{rgb}{0.0, 0.0, 0.0}
\definecolor{white}{rgb}{1.0, 1.0, 1.0}
\definecolor{darklime}{RGB}{102, 153, 0}
\definecolor{darkerlime}{RGB}{92, 123, 0}
\definecolor{evendarkerlime}{RGB}{67, 89, 0}

\definecolor{mednext}{HTML}{7d2727}
\definecolor{swinunetr}{HTML}{404d2e}
\definecolor{transunet}{HTML}{3b3f47}

\definecolor{fblack}{rgb}{0.0, 0.0, 0.0}
\definecolor{fblue}{rgb}{0.12156862745098039, 0.4666666666666667, 0.7058823529411765}
\definecolor{fred}{rgb}{0.8392156862745098, 0.15294117647058825, 0.1568627450980392}
\definecolor{fgreen}{rgb}{0.17254901960784313, 0.6274509803921569, 0.17254901960784313}
\definecolor{fcyan}{rgb}{0.09019607843137255, 0.7450980392156863, 0.8117647058823529}
\definecolor{fpurple}{rgb}{0.5803921568627451, 0.403921568627451, 0.7411764705882353}
\definecolor{fpink}{rgb}{0.8901960784313725, 0.4666666666666667, 0.7607843137254902}
\definecolor{fgray}{rgb}{0.4980392156862745, 0.4980392156862745, 0.4980392156862745}
\definecolor{forange}{rgb}{1.0, 0.4980392156862745, 0.054901960784313725}
\definecolor{fwhite}{rgb}{1, 1, 1}

\definecolor{lightblue}{RGB}{210, 233, 255}
\definecolor{lightyellow}{RGB}{222, 255, 209}

\newcommand{\JZR}[1]{}

\newcommand{\Jr}[1]{}

\begingroup\lccode`~=`_
\lowercase{\endgroup\def~}#1{_{\scriptscriptstyle#1}}

\AtBeginDocument{\mathcode`_="8000 \catcode`_=12 }

\makeatletter
\renewcommand{\@seccntformat}[1]{}
\newcommand{\sectionn}{\@startsection{section}{1}{0mm}%
                                {-\baselineskip}%
                                {0.5\baselineskip}%
                                {\normalfont\large\bfseries\MakeUppercase}}%
\makeatother

\newcommand\rurl[1]{%
  \href{http://#1}{\nolinkurl{#1}}%
}

\sectionfont{\fontsize{18}{18}\selectfont}

\soulregister\ref7
\soulregister\cite7
\soulregister\pageref7
\soulregister\citet7

\title{\bfseries A Visual Grounding Benchmark for Retinal OCT with Multimodal Large Language Models}

\author[1,*]{Ronald Fecso}
\author[1]{Jos\'e Morano}
\author[1]{Botond Fazekas}
\author[2]{Paul Widmann-Sedlnitzky}
\author[2]{Dorota Himel}
\author[1]{Laetita Hinterhuber}
\author[1]{Virginia Mares}
\author[2,*]{Andreas Pollreisz}
\author[1,*]{Hrvoje Bogunovi\'c}
\affil[1]{\small Institute of Artificial Intelligence, Center for Medical Data Science, Medical University of Vienna, Vienna, Austria}
\affil[2]{Department of Ophthalmology and Optometry, Medical University of Vienna, Vienna, Austria}
\affil[*]{\texttt{ronald.fecso@meduniwien.ac.at, andreas.pollreisz@meduniwien.ac.at, hrvoje.bogunovic@meduniwien.ac.at}}

\date{}  %

\begin{document}

\maketitle

\renewcommand{\figurename}{Fig.}

\begin{abstract}
Multimodal large language models (MLLMs) can answer clinical questions about medical images, but trustworthy use requires visual grounding: models should localize the pathology they report rather than rely on spurious shortcuts. Whether MLLMs can visually ground clinical findings in retinal optical coherence tomography (OCT), the most widely used imaging modality in ophthalmology, remains largely unexplored. To address this gap, we introduce RetinaGroundBench, the first benchmark for biomarker visual grounding in retinal OCT. It comprises 815 B-scans exhaustively annotated by ophthalmologists with 4,537 bounding boxes covering 13 biomarkers across 8 diseases. Because MLLMs often fragment a single lesion into multiple partial predictions, we propose AggIoU, a metric that evaluates whether the union of overlapping predicted boxes covers the annotated region. Across 27 open-source MLLMs (0.8--38B parameters), we find that visual grounding remains largely unsolved: the best general-purpose model achieves a macro F1 of just 23.7\%, with consistently low recall, despite reaching up to 73\% balanced accuracy on biomarker presence questions. To narrow this gap, we developed RetGrounder, an MLLM fine-tuned using an OCT-specific strategy integrating grounding and textual information. Despite its relatively small size (8B), RetGrounder improves grounding F1 to 33.9\%, a 43\% relative improvement. Our benchmark provides a reference point for measuring progress in the field of OCT diagnosis with MLLMs.
\end{abstract}

\section{Introduction}

Multimodal Large Language Models (MLLMs), which combine language and vision encoders to jointly process images and text, have demonstrated great potential as assistants to ophthalmologists~\cite{Abbas2026, Carla2025-ee, yang2024advancingmultimodalmedicalcapabilities, Zhang2024, Agbareia2025-zm,Holland2025}. The visual question-answering (VQA) ability of these models is of particular interest, as it allows users to query a model about the presence of diseases or lesions in an image~\cite{Tran2025}. However, for such assistants to be trusted in clinical practice, they must not only answer accurately, but also indicate the visual evidence behind each answer.
This capability is known as \emph{visual grounding}, in which a model outputs the spatial location of a queried finding.
Without such grounding, a textual answer gives clinicians no way to confirm that a model attended to the relevant pathology rather than exploiting spurious correlations or learned shortcuts~\cite{DeGrave2021}, or simply hallucinating. This lack of verifiability is a central barrier to clinical adoption~\cite{Kelly2019}. Assessing whether current MLLMs can be trusted therefore requires understanding not only their diagnostic question answering ability but also their capacity to spatially localise the pathology underlying those answers.

Most previous studies focused on evaluating MLLMs in terms of
examination performance and diagnosis on real clinical cases~\cite{Abbas2026, Carla2025-ee, yang2024advancingmultimodalmedicalcapabilities, Zhang2024, Agbareia2025-zm}. In the former category, frontier models (including 
GPT-5~\cite{gpt5_2025}, Gemini 2.5 Pro~\cite{gemini25_2025}, Claude Sonnet 4.0~\cite{claudesonnet4_2025}, and DeepSeek-V3~\cite{deepseekv3_2024}) have improved substantially on the Fellowship of the Royal College of Ophthalmologists (FRCOphth) examinations since 2023, with the best-performing model, Gemini 2.5 Pro, exceeding typical top human candidate scores. However, error analysis still revealed critical reasoning deficits such as premature diagnostic closure and confidently unjustified decisions~\cite{Abbas2026}. On real clinical cases, rather than curated examination questions, the evidence is more mixed: on 50 retinal optical coherence tomography (OCT) and OCT angiography (OCT-A) cases, GPT-4o diagnosed 62\% correctly versus Gemini's 32\%~\cite{Carla2025-ee}; Med-Gemini surpassed baselines on 18 of 20 medical image classification tasks including ophthalmology~\cite{yang2024advancingmultimodalmedicalcapabilities}; and GPT-4o matched clinicians in differential glaucoma diagnosis on OCT while struggling with primary diagnosis~\cite{Zhang2024}, with further evidence of effective few-shot retinal disease diagnosis on OCT~\cite{Agbareia2025-zm}.

Several models have also been developed specifically for ophthalmology. OphGLM~\cite{DENG2024103001} was the first, combining ConvNext~\cite{liu2022convnet2020s}, BERT~\cite{devlin-etal-2019-bert}, and ChatGLM-6B~\cite{glm2024chatglm} for fundus diagnostics; VisionUnite~\cite{11124413} added open-ended multi-disease fundus diagnosis with clinical explanations, and EyeCLIP~\cite{Shi2025}, pre-trained on 2.77 million images across 11 modalities, achieved strong classification and cross-modal retrieval. For OCT specifically, LO-VLM~\cite{Haghighi2025} targets anatomical layer-specific interpretation and RetinaVLM~\cite{Holland2025}, age-related macula degeneration (AMD) staging,
approaching junior-ophthalmologist performance. Despite this diversity, these models share a limitation that matters for clinical verification: their outputs are text---a diagnosis, a disease stage, or a free-text description---with no indication of the image evidence behind them. Their scope is narrow as well, and the two OCT-specific models each address a single task. None can be queried for the location of an arbitrary biomarker.

Several ophthalmology-specific VQA benchmarks have been proposed, though none assess visual grounding. LMOD+~\cite{qin2026lmodcomprehensivemultimodaldataset} is the most systematic, testing 24 MLLMs on 32,633 images across 5 ophthalmic imaging modalities including OCT, but its spatial annotations cover only anatomical structures (e.g.,\ retina, choroid), not fine-grained biomarkers. OphthalWeChat~\cite{XU202633} and MedBench~\cite{liang2025novelophthalmicbenchmarkevaluating} likewise evaluate ophthalmic VQA, and FunBench~\cite{WeiQij_FunBench_MICCAI2025} organises tasks hierarchically across modality, anatomy, lesion, and disease. Across all of these, the performance of MLLMs still lags behind supervised networks, and no benchmark requires models to localise the pathology underlying their answers.

To contextualize existing gaps, prior work can be placed along two hierarchies. The image domain narrows from general medical imaging, through ophthalmology, to retinal OCT specifically; the task deepens from language-only assessment, through VQA on an image, to grounded VQA, in which a model must also localise the finding it reports. Retinal OCT already appears at the shallow end of the task hierarchy---evaluated, in the benchmarks above, as VQA or anatomical recognition, but never biomarker grounding. Grounding, conversely, has been explored further in other specialties---radiology~\cite{boecking}, pathology~\cite{10.1007/978-3-032-05169-1_44}, and sequential medical images~\cite{yue2025medsgbench}---but not brought down to ophthalmic biomarkers. No prior benchmark occupies the intersection of the two: grounded VQA of biomarkers on retinal OCT, despite it being the most widely used imaging modality in ophthalmology and the diagnostic gold standard for a broad range of retinal diseases.

\begin{figure}[p]
    \centering
    \includegraphics[width=\linewidth]{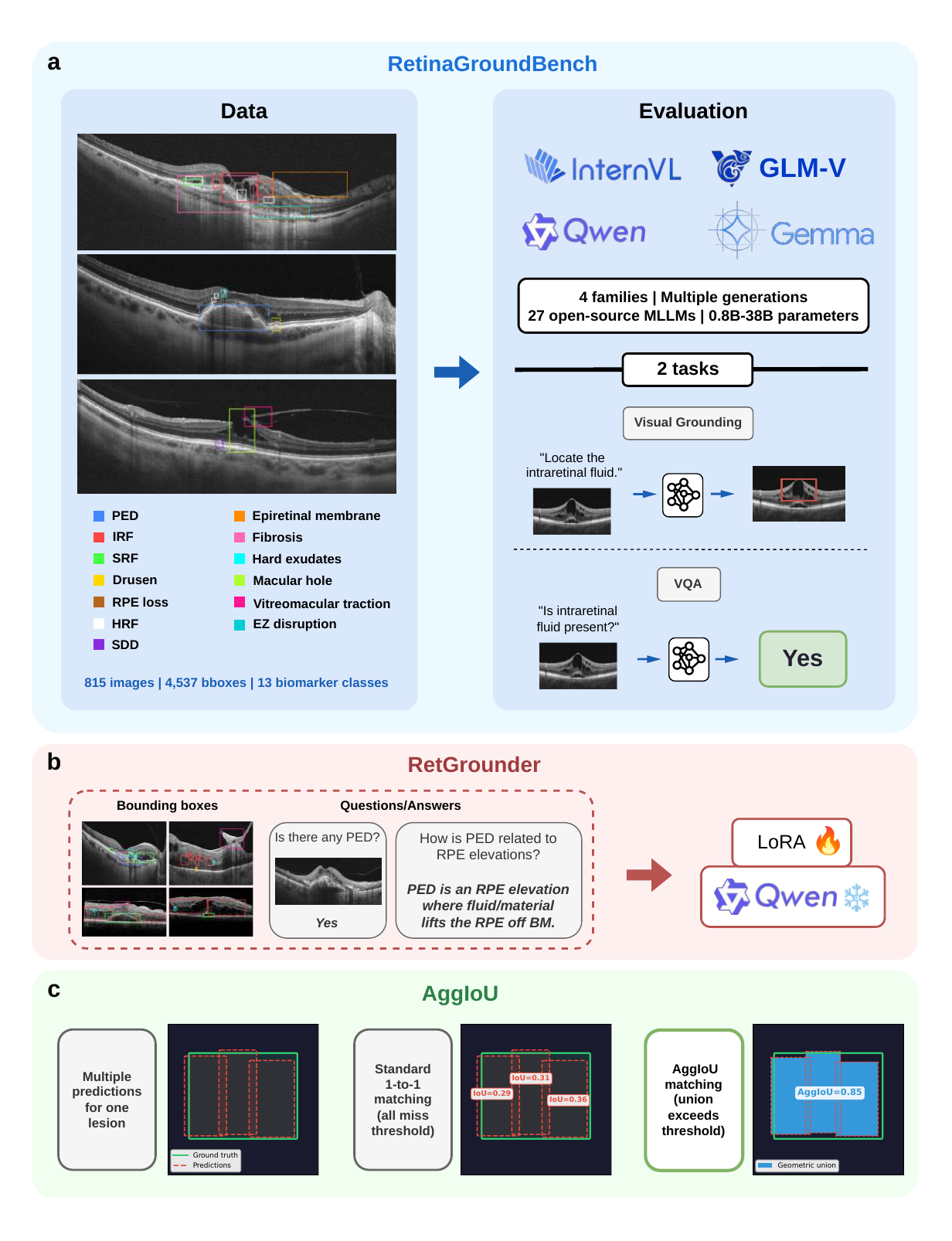}
    \caption{\textbf{Contributions overview.} 
    \textbf{a.} RetinaGroundBench benchmark. We evaluate 27 state-of-the-art open-source MLLMs spanning 4 model families (0.8B--38B parameters) on visual grounding and VQA for the localisation of 13 biomarkers in retinal OCT.
    \textbf{b.} RetGrounder model trained using LoRA on a dataset with bounding boxes of lesions and question--answer pairs.
    \textbf{c.} AggIoU matching strategy. Standard one-to-one matching evaluates each prediction independently, making all of them fall below the IoU threshold and thus counting the lesion as a false negative (Precision/Recall = 0.33/1.0). In contrast, AggIoU computes the geometric union of gathered predictions, yielding a high overlap that correctly identifies the lesion as a true positive (Precision/Recall = 1.0/1.0).
    }
    \label{fig:graphical_abstract}
\end{figure}

Purely in terms of data, retinal OCT is well served by classification, segmentation, and lesion detection datasets~\cite{Bogunovic2019-vr, Kermany2018, prabhushankarolives2022, Kulyabin2024, Morano2025, Arikan2025}, but none supports the task we study. Classification datasets carry no spatial annotation. Segmentation and detection datasets do localise lesions, but their annotations are tied to a closed set of classes fixed in advance rather than to a clinician's query, so they cannot test whether a model retrieves the correct finding when asked in natural language; and their dense-mask output is mismatched to the text coordinates MLLMs emit, so scoring an MLLM against them would measure mask generation rather than spatial grounding. 

Grounding retinal OCT biomarkers is not simply the same task in a new modality; three obstacles stand in the way. The first is the pathology. Unlike the discrete, well-bounded objects of natural-image grounding, many retinal biomarkers are diffuse and low-contrast, recur as many small instances across a single B-scan, and have extents expert graders themselves would draw differently, e.g., drusen and 
fibrosis.
The second is the output format: an MLLM must serialise a spatial extent as coordinate tokens through an autoregressive language decoder, so whatever spatial detail its vision encoder captures must survive translation into text, a constraint no purpose-built detector faces. The third is measurement: conventional detection metrics assume one box per instance and match greedily one-to-one, yet MLLMs scatter several partial boxes across a single lesion, so a model that has in substance found the lesion is scored as one hit and several false alarms. Any assessment of medical grounding must therefore contend not only with whether models can localise pathology, but with whether our instruments can tell.

These obstacles, together with the absence of ophthalmic visual grounding benchmarks and the clinician's need to verify that model answers are visually grounded, motivate our benchmark contribution, RetinaGroundBench.
This is the first visual grounding benchmark for retinal OCT lesion localisation, spanning 13 lesion types evaluated across state-of-the-art open-source MLLMs. To measure grounding as MLLMs actually perform it, we further propose Aggregated IoU (AggIoU) (Fig.~\ref{fig:graphical_abstract}c), which scores each ground-truth box against the \emph{union} of the predictions that overlap it rather than against a single greedily matched box: a lesion covered collectively by several partial predictions is credited as found, while predictions that substantially overshoot or undershoot the target are still penalised. Finally, motivated by the documented failure of out-of-the-box MLLMs at spatial localisation for OCT, we introduce RetGrounder, a domain-adapted model demonstrating that diverse targeted fine-tuning can substantially narrow the medical visual grounding gap. In summary, this work contributes a new benchmark (RetinaGroundBench), a new visual grounding metric (AggIoU), and a new domain-adapted MLLM (RetGrounder) capable of visual grounding on retinal OCT.

\section{Results}

\subsection{Overview}
Fig.~\ref{fig:graphical_abstract} gives an overview of the benchmark and the evaluation framework. RetinaGroundBench comprises 815 retinal OCT B-scans drawn from two public datasets: OCTDL~\cite{Kulyabin2024} and OLIVES~\cite{prabhushankarolives2022}, annotated by ophthalmologists with 4{,}537 bounding boxes across 13 biomarker classes (Methods, Table~\ref{tab:gt_distribution}):
drusen, ellipsoid zone (EZ) disruption, epiretinal membrane (ERM), fibrosis, hard exudates (HE), hyperreflective foci (HRF), intraretinal fluid (IRF), macular hole (MH), pigment epithelial detachment (PED), retinal pigment epithelium (RPE) loss, subretinal drusenoid deposits (SDD), subretinal fluid (SRF), and vitreomacular traction (VMT).

Table~\ref{tab:dataset_comparison} compares RetinaGroundBench with existing retinal OCT datasets that provide spatial annotations and with ophthalmic MLLM benchmarks. OCT5k supplies biomarker bounding boxes but no MLLM protocol, and although its class count is comparable (9), its categories are concentrated within AMD-related biomarkers (e.g.\ five drusen subtypes), whereas RetinaGroundBench spans 13 distinct pathology types across 8 diseases. Conversely, ophthalmic MLLM benchmarks evaluate multimodal models but do not target biomarker grounding: LMOD+ restricts its spatial annotations to anatomical structures rather than fine-grained lesions, while FunBench and OphthalWeChat are VQA-only and carry no bounding-box annotations at all. RetinaGroundBench is thus the only resource that pairs biomarker-level bounding boxes with both a grounding and a VQA protocol, and it does so over more diseases and more biomarker classes than any of the alternatives, and across more evaluated models. Its image count is deliberately smaller: every box was placed by hand by retinal specialists, and the benchmark is sized for exhaustive per-instance annotation of all 13 biomarkers in every scan rather than for scale.
Unlike previous datasets, these classes are not a curated subset of lesions of interest, but the complete set of pathological findings observed in each image, so that the ground truth reflects a comprehensive assessment of every scan.

\begin{table}[btp]
\centering
\caption{\textbf{Comparison of RetinaGroundBench with existing retinal OCT detection datasets and ophthalmic MLLM benchmarks.} Biomarker VG = text-prompted visual grounding of pathological biomarkers; VQA = visual question answering; Dis.\ = diseases covered; BBox = bounding boxes. $^\dagger$Multi-modality benchmark: for LMOD+ the OCT subset is shown, whereas FunBench (fundus) and OphthalWeChat (29 modalities) are shown in full. RetinaGroundBench is the only resource combining biomarker-level bounding boxes with both a grounding and a VQA protocol for MLLMs, and covers the widest range of diseases and biomarker classes.}\label{tab:dataset_comparison}
\small
\resizebox{\textwidth}{!}{%
\begin{tabular}{@{}l cc cc l cc@{}}
\toprule
& \multicolumn{2}{c}{\textbf{MLLM protocol}} & \multicolumn{2}{c}{\textbf{Coverage}} & & & \\
\cmidrule(lr){2-3}\cmidrule(lr){4-5}
\textbf{Dataset} & \textbf{Biom. VG} & \textbf{VQA} & \textbf{Dis.} & \textbf{Cls.} & \textbf{Annot.} & \textbf{Images} & \textbf{Models} \\
\midrule
OCT5k~\cite{Arikan2025} & No & No & 3 & 9 & 4{,}698 BBox & 566 & --- \\
LMOD+$^\dagger$~\cite{qin2026lmodcomprehensivemultimodaldataset} & No & Yes & 2 & 4 & 9{,}262 BBox (anat.) & 3{,}859 & 24 \\
FunBench$^\dagger$~\cite{WeiQij_FunBench_MICCAI2025} & No & Yes & --- & --- & Image-level & 16{,}348 & 10 \\
OphthalWeChat$^\dagger$~\cite{XU202633} & No & Yes & --- & --- & Image-level & 3{,}469 & 6 \\
\midrule
\textbf{RetinaGroundBench} & \textbf{Yes} & \textbf{Yes} & \textbf{8} & \textbf{13} & \textbf{4{,}537 BBox} & \textbf{815} & \textbf{28} \\
\bottomrule
\end{tabular}%
}
\end{table}

In our benchmark, we evaluate 27 state-of-the-art, general-purpose open-source MLLMs spanning four model families and multiple generations: Qwen3-VL, Qwen3.5, Qwen3.6, InternVL3, InternVL3.5, GLM-4.6V-Flash, and Gemma-4, with parameter counts ranging from 0.8B to 38B (see full list in Supplementary Table~\ref{tab:model_repos}). All models were selected for their open weights and native bounding box output capability.

Two additional components of this framework are also introduced in this work. %
The first is RetGrounder, obtained by fine-tuning Qwen3-VL-8B-Instruct on a mixture of retinal OCT data spanning two task types: grounding data, comprising an in-house set of 34{,}112 annotated B-scans together with four public segmentation and detection datasets converted to bounding-box labels; and question-answering (QA) data, including classification data recast into VQA form and text-only QA pairs covering OCT biomarker knowledge.
Our model was also evaluated on RetinaGroundBench, resulting in 28 total benchmarked models.

The second is AggIoU (Aggregated IoU), our grounding metric.
MLLMs rarely place a single box neatly around a lesion; instead, they tend to scatter several smaller boxes across it (Fig.~\ref{fig:graphical_abstract}c), whereas human annotators often draw relatively large boxes spanning multiple instances of the same lesion~\cite{lepetit2026retinal}.
One-to-one detection matching accepts one of those boxes as the match and charges the rest as false positives, penalising a model that has in fact found the lesion. AggIoU instead scores each ground-truth box against the \emph{union} of the predictions that overlap it: predictions overlapping a ground-truth box by more than a fraction
of the smaller box's area are collected and their exact geometric union is recorded. AggIoU is then calculated as the IoU between the ground-truth box and that union. A ground-truth box counts as detected when its AggIoU exceeds a predefined value.
The metric therefore plays a dual role---it is both the reported measure of localisation quality and the criterion deciding true positives---and it credits predictions that collectively cover a lesion while still penalising those that are far larger or smaller than the target.
The Methods section describes in detail
the full procedure.

Each model is assessed on two axes: visual grounding and visual question answering (VQA). For grounding, we report mean AggIoU alongside precision, recall, and F1; for VQA, we report balanced accuracy on binary presence questions. All metrics are computed over the combined OCTDL and OLIVES evaluation set: per-class scores are obtained by pooling all images from both datasets, then macro-averaged across all 13 biomarker classes so that each class contributes equally regardless of its prevalence or source dataset.

\subsection{Visual grounding performance}

\begin{figure}[tbp]
    \centering
    \includegraphics[width=0.65\linewidth]{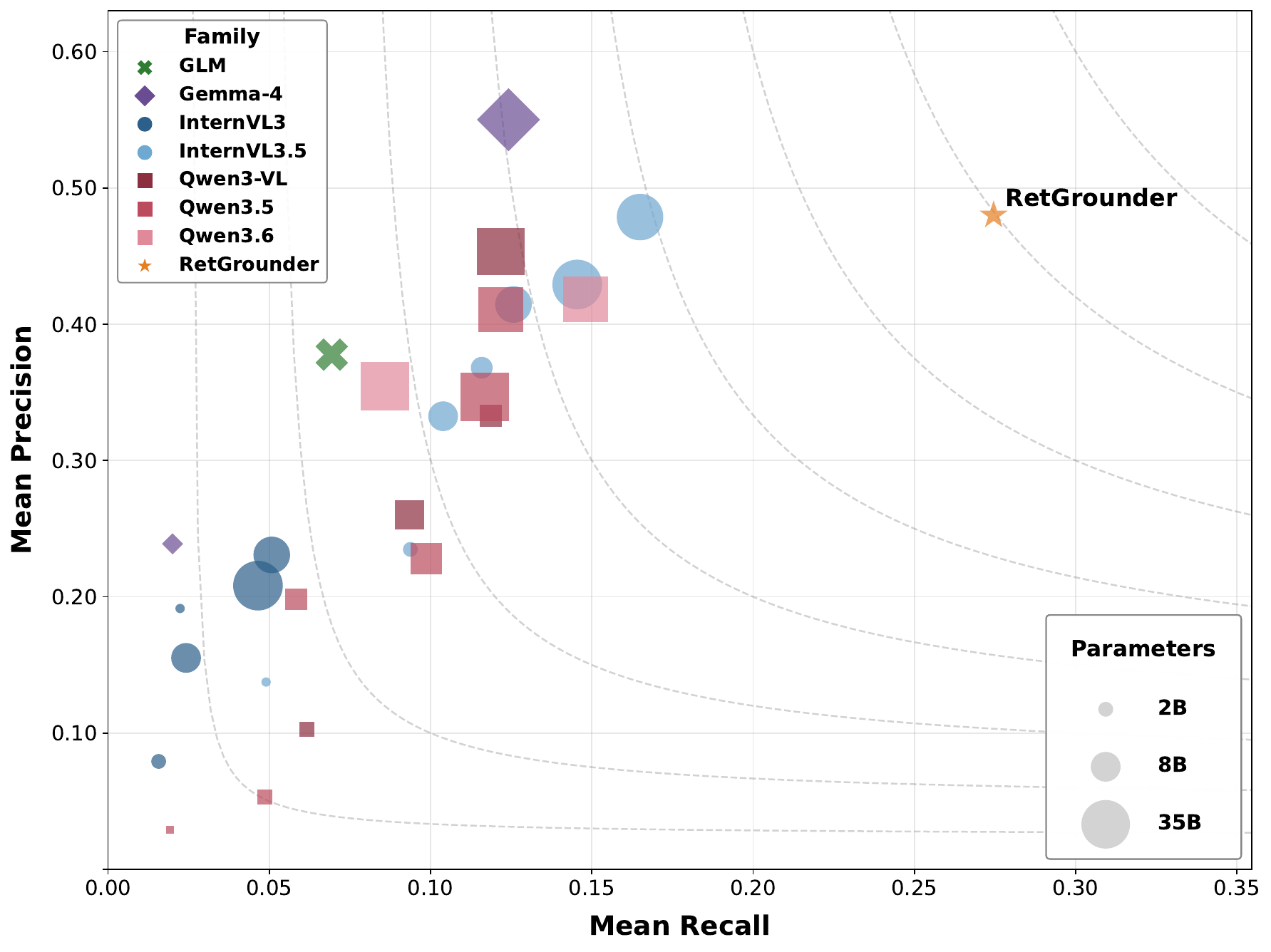}
    \caption{\textbf{Grounding precision vs.\ recall for all evaluated models.} Bubble size is proportional to model parameter count. Dashed lines indicate iso-F1 contours.
    }
    \label{fig:pr_bubble}
\end{figure}

Fig.~\ref{fig:pr_bubble} places the performance of 27 general-purpose models and RetGrounder in the precision--recall plane, with bubble size proportional to parameter count and dashed iso-F1 contours for reference.
Complementarily, Table~\ref{tab:vg_eval_custom} reports mean AggIoU, F1, precision, and recall for the same models.

\begin{table}[tbph]
\centering
\caption{\textbf{Visual grounding evaluation across models.} All metrics are macro-averaged across biomarker classes ($\pm$ standard deviation across classes).
Significance between best and second-best via paired Wilcoxon signed-rank test over GT bounding boxes:
$^{***}p<0.001$.}
\label{tab:vg_eval_custom}
\resizebox{0.9\textwidth}{!}{%
\begin{tabular}{llllll}
\toprule
\textbf{Model} & \textbf{Size} & \textbf{AggIoU} & \textbf{F1} & \textbf{Precision} & \textbf{Recall} \\
\midrule
\multirow{4}{*}{Qwen3-VL}
& 2B                     & 5.9$\pm$4.1 & 6.3$\pm$4.8 & 10.3$\pm$10.3 & 6.2$\pm$4.2 \\
& 4B                     & 9.2$\pm$8.2 & 16.9$\pm$17.8 & 33.3$\pm$32.1 & 11.9$\pm$12.6 \\
& 8B                     & 7.3$\pm$6.0 & 13.3$\pm$13.7 & 26.0$\pm$25.8 & 9.3$\pm$9.4 \\
& 32B                    & 9.5$\pm$7.0 & 18.8$\pm$15.6 & 45.3$\pm$33.3 & 12.2$\pm$10.4 \\
\cmidrule{1-6}
\multirow{6}{*}{Qwen3.5}
& 0.8B                   & 2.0$\pm$1.4 & 1.6$\pm$1.3 & 2.9$\pm$5.2 & 1.9$\pm$1.5 \\
& 2B                     & 4.2$\pm$5.1 & 4.0$\pm$3.9 & 5.3$\pm$5.4 & 4.9$\pm$6.2 \\
& 4B                     & 6.0$\pm$5.4 & 8.8$\pm$9.4 & 19.8$\pm$25.1 & 5.8$\pm$5.8 \\
& 9B                     & 8.1$\pm$6.4 & 13.0$\pm$13.7 & 22.8$\pm$27.1 & 9.9$\pm$9.1 \\
& 27B                    & 8.9$\pm$6.9 & 18.2$\pm$18.1 & 41.1$\pm$33.1 & 12.2$\pm$12.8 \\
& 35B-A3B                & 9.6$\pm$6.8 & 16.6$\pm$15.4 & 34.7$\pm$35.5 & 11.7$\pm$9.9 \\
\cmidrule{1-6}
\multirow{2}{*}{Qwen3.6}
& 27B                    & 10.8$\pm$9.0 & 21.0$\pm$21.7 & 41.8$\pm$35.8 & 14.8$\pm$16.7 \\
& 35B-A3B                & 8.1$\pm$6.4 & 13.3$\pm$13.3 & 35.4$\pm$32.2 & 8.6$\pm$8.8 \\
\cmidrule{1-6}
\multirow{1}{*}{GLM-4.6V-Flash}
& 9B                     & 7.1$\pm$4.5 & 11.5$\pm$10.5 & 37.8$\pm$33.6 & 6.9$\pm$6.4 \\
\cmidrule{1-6}
\multirow{5}{*}{InternVL3}
& 1B                     & 3.5$\pm$3.0 & 4.0$\pm$4.8 & 19.1$\pm$23.2 & 2.2$\pm$2.7 \\
& 2B                     & 2.5$\pm$2.2 & 2.6$\pm$3.5 & 7.9$\pm$9.7 & 1.6$\pm$2.2 \\
& 8B                     & 3.0$\pm$2.7 & 4.1$\pm$6.3 & 15.5$\pm$18.9 & 2.4$\pm$3.8 \\
& 14B                    & 5.1$\pm$4.1 & 8.2$\pm$8.1 & 23.1$\pm$19.6 & 5.1$\pm$5.2 \\
& 38B                    & 5.2$\pm$5.0 & 7.4$\pm$9.2 & 20.8$\pm$21.7 & 4.6$\pm$6.1 \\
\cmidrule{1-6}
\multirow{7}{*}{InternVL3.5}
& 1B                     & 4.6$\pm$4.5 & 7.1$\pm$8.7 & 13.7$\pm$16.6 & 4.9$\pm$6.1 \\
& 2B                     & 7.9$\pm$7.1 & 13.1$\pm$15.2 & 23.5$\pm$26.5 & 9.4$\pm$11.1 \\
& 4B                     & 9.3$\pm$8.3 & 16.7$\pm$19.1 & 36.8$\pm$29.6 & 11.6$\pm$15.2 \\
& 8B                     & 8.6$\pm$6.6 & 15.4$\pm$16.1 & 33.2$\pm$29.0 & 10.4$\pm$11.5 \\
& 14B                    & 10.4$\pm$7.2 & 18.8$\pm$16.4 & 41.4$\pm$29.5 & 12.6$\pm$11.5 \\
& 30B-A3B                & \underline{12.6$\pm$9.7} & \underline{23.7$\pm$20.6} & 47.9$\pm$29.1 & \underline{16.5$\pm$15.9} \\
& 38B                    & 10.8$\pm$8.3 & 21.1$\pm$19.5 & 42.9$\pm$32.2 & 14.5$\pm$14.6 \\
\cmidrule{1-6}
\multirow{2}{*}{Gemma-4}
& E2B                    & 1.4$\pm$3.2 & 3.3$\pm$9.2 & 23.9$\pm$34.8 & 2.0$\pm$5.8 \\
& 26B-A4B                & 8.0$\pm$8.9 & 18.5$\pm$20.8 & \textbf{55.0$\pm$32.6}$^{***}$ & 12.4$\pm$16.8 \\
\midrule
RetGrounder & 8B & \textbf{18.5$\pm$13.9}$^{***}$ & \textbf{33.9$\pm$26.2}$^{***}$ & \underline{48.0$\pm$32.4} & \textbf{27.5$\pm$23.7}$^{***}$ \\
\bottomrule
\end{tabular}%
}
\end{table}

The resulting landscape is markedly asymmetric: all evaluated models exhibited a pronounced precision--recall imbalance for visual grounding, with precision exceeding recall for every model. Most open-source models clustered in the low-recall region, with mean recall below 25\%. Gemma-4-26B-A4B achieved the highest precision (55.0\%) among general-purpose models but with low recall (12.4\%). InternVL3.5-30B-A3B was the best-performing general-purpose model overall (F1\,=\,23.7\%, precision\,=\,47.9\%, recall\,=\,16.5\%). RetGrounder achieved the highest recall (27.5\%) while maintaining competitive precision (48.0\%), resulting in a macro F1 of 33.9\%---a relative improvement of 43\% over InternVL3.5-30B-A3B, the second-best model. It is worth noting that
recall
is particularly consequential clinically:
a spurious box is visible to the reader and can be dismissed at a glance, whereas a biomarker the model never localises leaves no trace in the output for a clinician to check.
In automated or semi-automated setups, this would translate into patients with disease not being referred for further examination.
On the other hand, over-detection would cause unnecessary subsequent assessments, burdening clinical workflows.
Thus, while high recall is fundamental for patient outcomes, the precision of the model also represents a key measure of its applicability in real-world scenarios.
This balance is better achieved by 
RetGrounder.

\begin{figure}[btp]
    \centering
    \includegraphics[width=\linewidth]{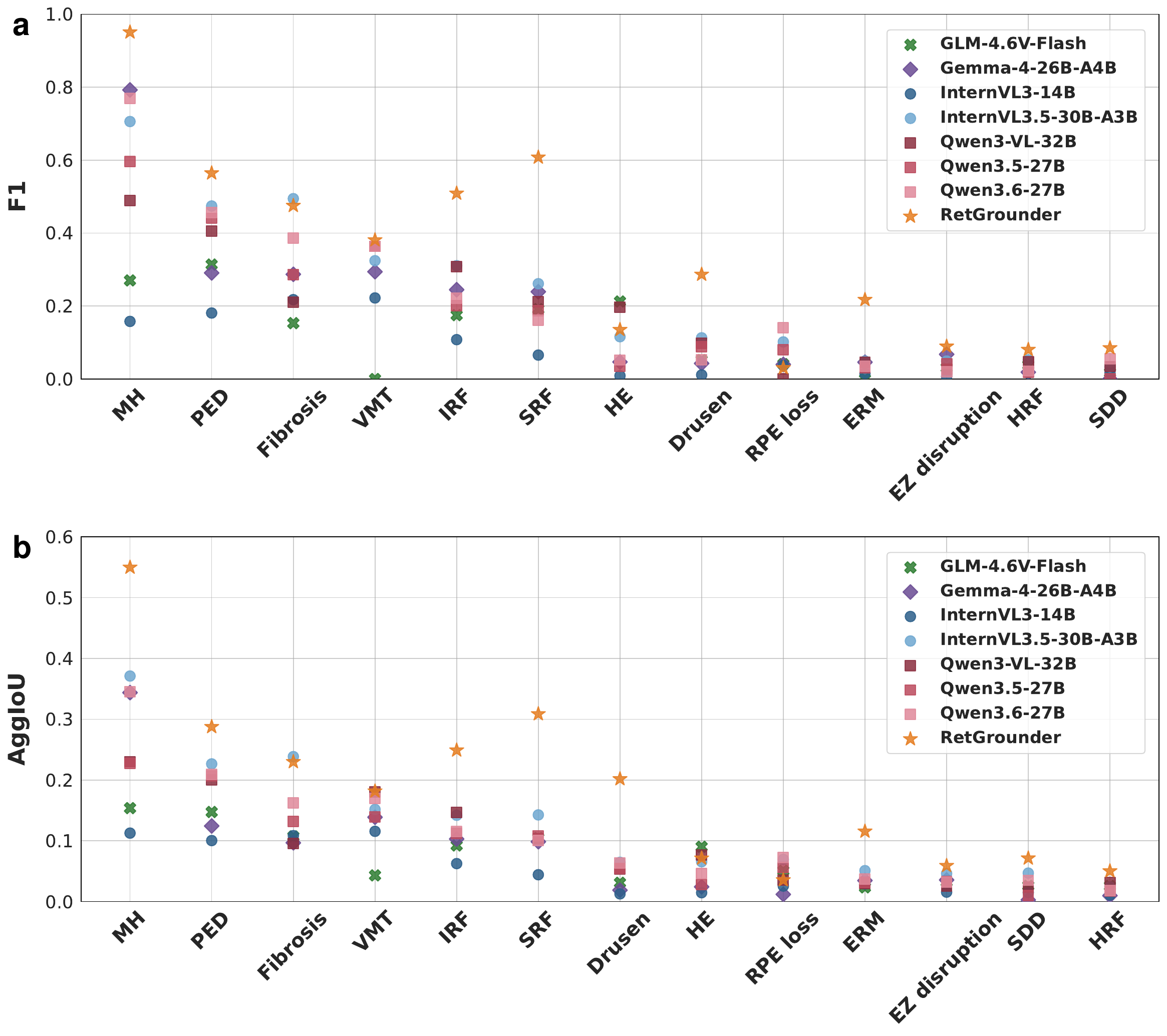}
    \caption{\textbf{Per-biomarker grounding results.}
    \textbf{a.} F1 for the best-performing model in each family--generation. Biomarkers are ordered by mean F1 (descending) across the displayed models.
    \textbf{b.} Mean AggIoU for the best-performing model in each family. Biomarkers are ordered by mean AggIoU (descending) across the displayed models.
    }
    \label{fig:biomarker_f1}
\end{figure}

Per-biomarker F1 scores (Fig.~\ref{fig:biomarker_f1}a; full per-class breakdowns for all models in Supplementary Tables~\ref{tab:perclass_aggiou}--\ref{tab:perclass_recall}) varied substantially across biomarker classes. Macular hole was localised most effectively (RetGrounder F1\,=\,95.1\%, InternVL3.5-30B-A3B F1\,=\,70.6\%). SRF (60.8\%), PED (56.4\%), and IRF (50.9\%) were localised with moderate success by RetGrounder. Hard exudates (F1\,=\,13.5\%), ERM (F1\,=\,21.7\%), and drusen (F1\,=\,28.7\%) proved difficult for all models. The four hardest classes---ellipsoid zone disruption, hyperreflective foci, subretinal drusenoid deposits, and retinal pigment epithelium loss---%
could not be detected reliably by any model, with even RetGrounder reaching only F1\,=\,8.9\%, 8.1\%, 8.5\%, and 3.0\%, respectively. Fibrosis and RPE loss were the only two classes on which RetGrounder failed to outperform InternVL3.5-30B-A3B, the strongest general-purpose model overall (F1\,=\,47.5\% vs.\ 49.4\% for fibrosis and 3.0\% vs.\ 10.2\% for RPE loss).

\begin{figure}[tbp]
    \centering
    \includegraphics[width=0.7\linewidth]{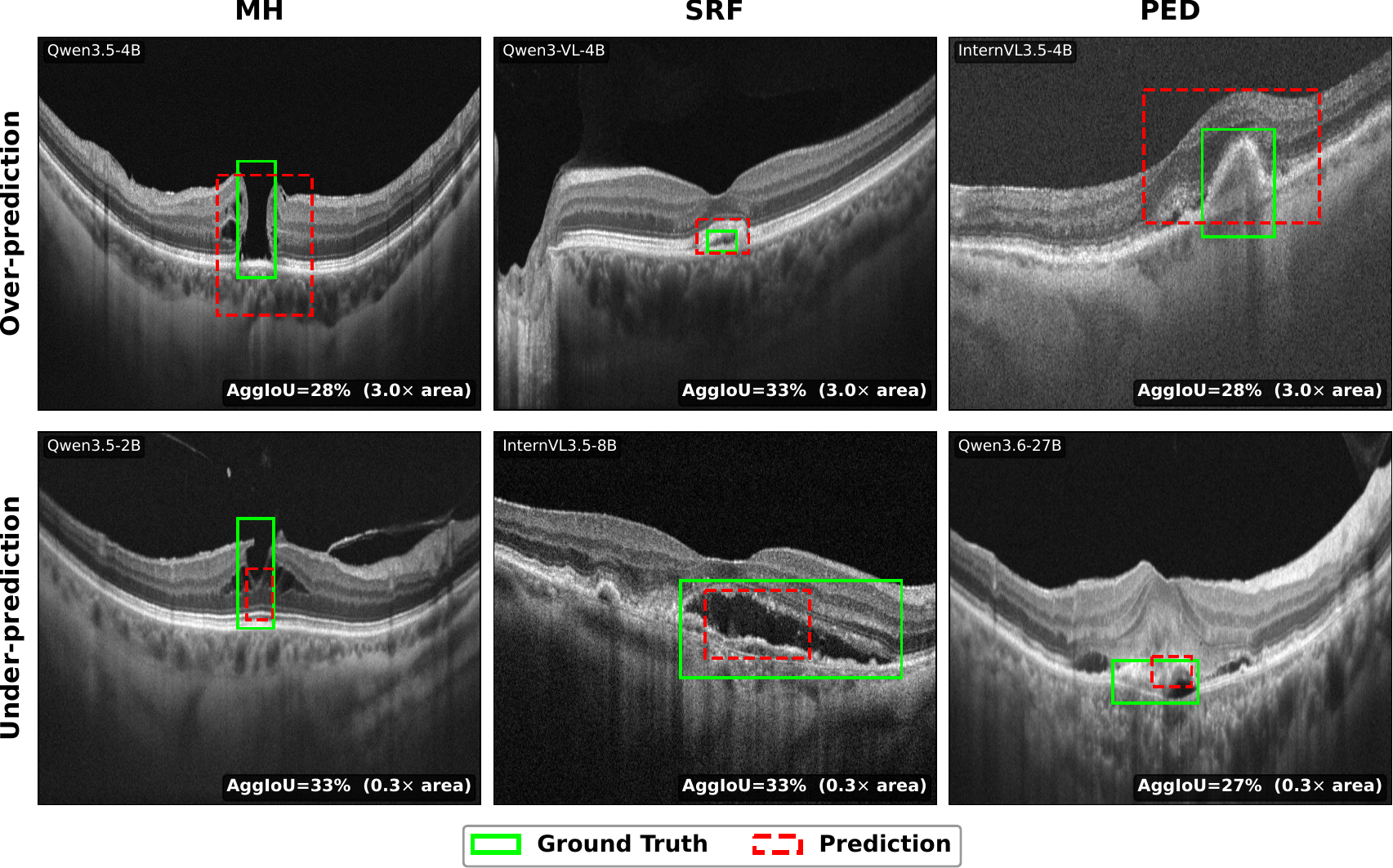}
    \caption{\textbf{Prediction size mismatch for general-purpose models.}
    Top row: over-prediction (predicted box larger than the ground truth); bottom row: under-prediction (predicted box smaller). Columns show macular hole (MH), subretinal fluid (SRF), and pigment epithelial detachment (PED). Green solid: ground truth; red dashed: prediction. Each panel is annotated with the per-image AggIoU and the predicted-to-ground-truth area ratio; the model providing each example is labelled top-left.}
    \label{fig:size_mismatch}
\end{figure}

Mean AggIoU scores (Fig.~\ref{fig:biomarker_f1}b) followed a similar pattern but remained universally low, reflecting poor spatial overlap even for detected biomarkers. RetGrounder achieved the highest AggIoU on macular hole (55.0\%), the only class where any model exceeded 50\%. SRF (30.9\%), PED (28.8\%), and IRF (24.9\%) had moderate overlap, while hard exudates (7.1\%) and ERM (11.6\%) remained near floor level. The gap between high F1 and low AggIoU for several classes---such as macular hole, where RetGrounder achieved F1\,=\,95.1\% but AggIoU\,=\,55.0\%---indicates that while models can identify the correct region, their predicted boxes are often substantially larger or smaller than the ground truth (Fig.~\ref{fig:size_mismatch}).

\begin{figure}[tbp]
    \centering
    \includegraphics[width=\linewidth]{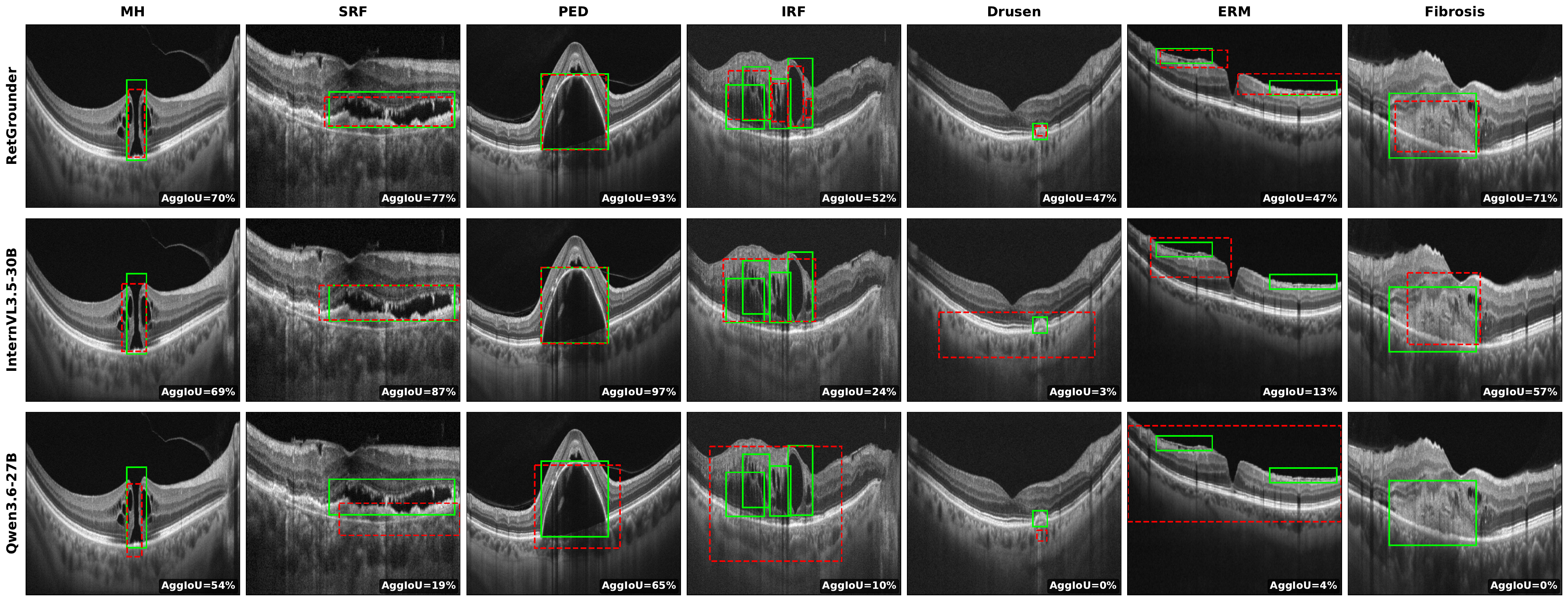}\\
    \includegraphics[width=\linewidth]{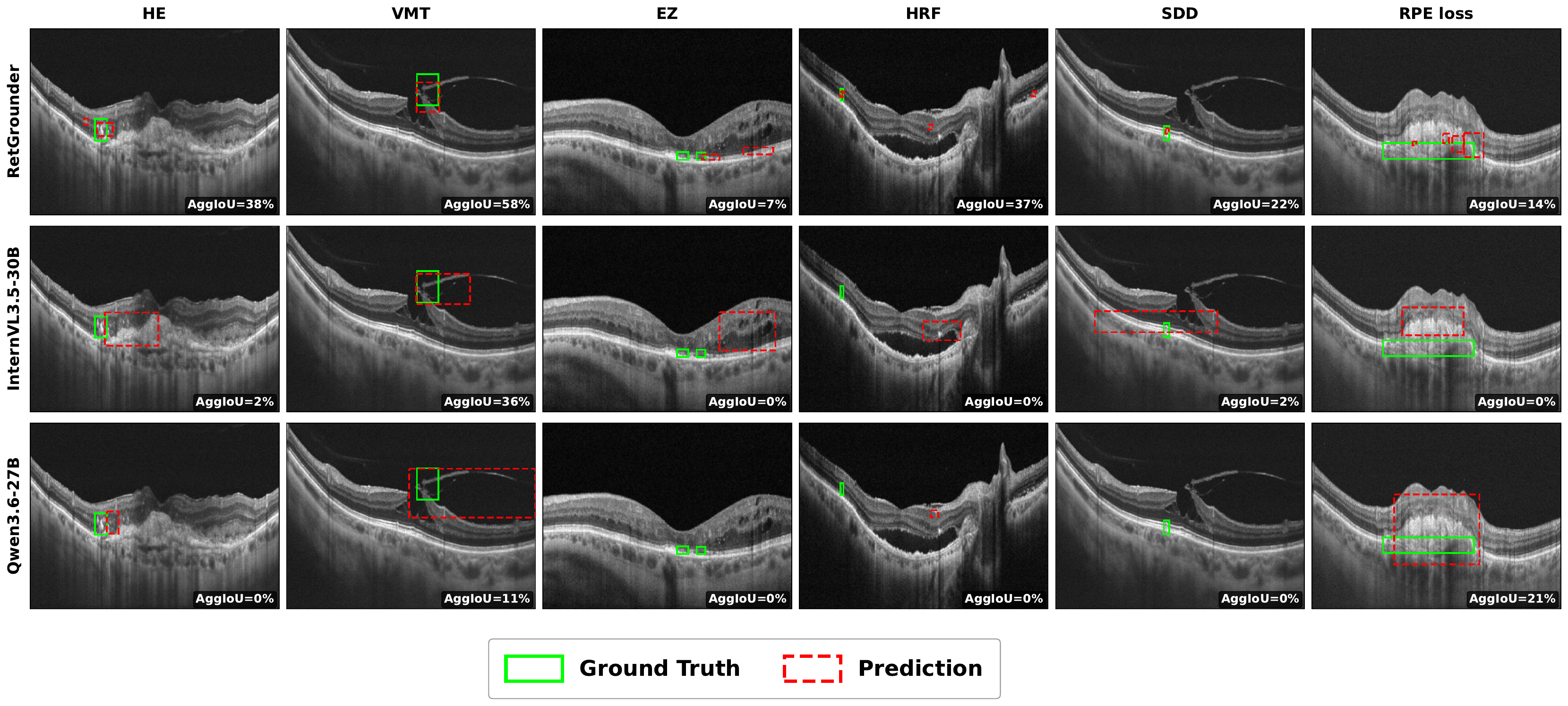}
    \caption{\textbf{Qualitative grounding examples for the three top-performing models.} Lesions include macular hole, SRF, PED, IRF, drusen, ERM, fibrosis, hard exudates, and vitreomacular traction. Green boxes: ground truth; red dashed boxes: predictions. Per-image AggIoU shown in each cell.}
    \label{fig:qualitative1}
\end{figure}

Fig.~\ref{fig:qualitative1}
shows representative grounding predictions from the three top-performing models across all 13 biomarker classes, with per-image AggIoU scores overlaid on each cell. RetGrounder produced predictions that more closely matched the ground truth extent, particularly for SRF, IRF, and macular hole. InternVL3.5-30B-A3B and Qwen3.6-27B tended to produce fewer or less precisely localised boxes, with predictions sometimes covering only a portion of the annotated region or missing it entirely. Hard exudates and vitreomacular traction proved challenging for all models, with AggIoU scores near zero for most predictions.

\subsection{VQA performance}

\begin{figure}[tbp]
    \centering
    \includegraphics[width=0.65\linewidth]{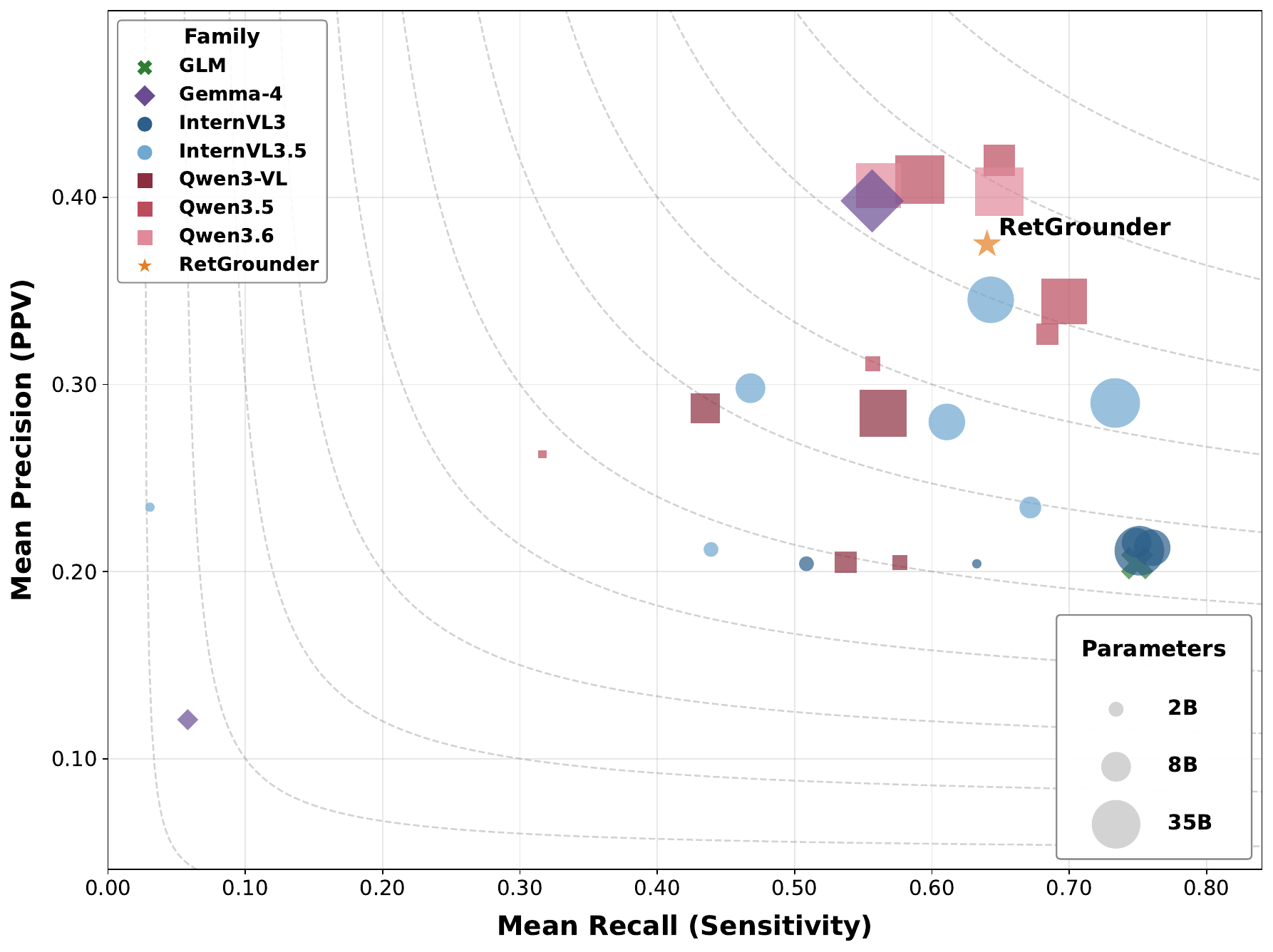}
    \caption{\textbf{VQA precision vs.\ recall for all evaluated models.} Compared to grounding (Fig.~\ref{fig:pr_bubble}), models are distributed across a much wider and higher-performing region.}
    \label{fig:vqa_pr_bubble}
\end{figure}

Fig.~\ref{fig:vqa_pr_bubble} shows the precision--recall landscape for VQA, and
Table~\ref{tab:vqa_eval} reports balanced accuracy, F1, precision, and recall on the binary presence questions for every evaluated model.

The VQA precision--recall landscape differed markedly from grounding. Models were distributed across a much wider and higher-performing region, with recalls reaching 76\% and macro F1 scores up to 49\%. Qwen3.5-9B (F1\,=\,48.8\%) and Qwen3.6-35B-A3B (F1\,=\,47.4\%) led VQA performance, while RetGrounder, despite dominating grounding, ranked among the stronger models without leading them (F1\,=\,40.3\%). Several models exhibited high recall but low precision (e.g.\ GLM-4.6V-Flash: recall\,=\,74.9\%, precision\,=\,20.4\%).

\begin{table}[tbp]
\centering
\caption{\textbf{Visual question answering evaluation.} All metrics are macro-averaged across all 13 biomarker classes ($\pm$ standard deviation across classes). Significance between the best and second-best model on balanced accuracy was assessed with a two-sided paired Wilcoxon signed-rank test over per-query correctness (one binary outcome per image--biomarker presence question): $^{***}p<0.001$.
}
\label{tab:vqa_eval}
\resizebox{0.9\textwidth}{!}{%
\begin{tabular}{llllll}
\toprule
\textbf{Model} & \textbf{Size} & \textbf{BAcc} & \textbf{F1} & \textbf{Precision} & \textbf{Recall} \\
\midrule
\multirow{4}{*}{Qwen3-VL}
& 2B                     & 55.7$\pm$8.0 & 26.7$\pm$18.9 & 20.5$\pm$16.0 & 57.7$\pm$32.2 \\
& 4B                     & 53.4$\pm$5.3 & 23.4$\pm$19.6 & 20.5$\pm$16.7 & 53.7$\pm$40.6 \\
& 8B                     & 56.8$\pm$9.4 & 22.8$\pm$20.1 & 28.7$\pm$22.9 & 43.5$\pm$37.9 \\
& 32B                    & 60.2$\pm$9.0 & 29.5$\pm$19.0 & 28.5$\pm$20.2 & 56.4$\pm$33.9 \\
\cmidrule{1-6}
\multirow{6}{*}{Qwen3.5}
& 0.8B                   & 56.0$\pm$5.5 & 23.0$\pm$17.6 & 26.3$\pm$22.5 & 31.6$\pm$23.9 \\
& 2B                     & 65.6$\pm$11.1 & 37.5$\pm$21.0 & 31.1$\pm$19.5 & 55.7$\pm$27.8 \\
& 4B                     & 68.2$\pm$11.2 & 40.2$\pm$18.1 & 32.7$\pm$16.9 & 68.4$\pm$25.4 \\
& 9B                     & \textbf{72.7$\pm$11.8}$^{***}$ & \textbf{48.8$\pm$19.1} & \textbf{42.0$\pm$20.1} & 64.9$\pm$19.5 \\
& 27B                    & 70.4$\pm$13.2 & 43.7$\pm$20.1 & 34.4$\pm$19.2 & 69.6$\pm$20.6 \\
& 35B-A3B                & 71.0$\pm$13.8 & 45.3$\pm$22.0 & \underline{40.9$\pm$22.0} & 59.1$\pm$26.2 \\
\cmidrule{1-6}
\multirow{2}{*}{Qwen3.6}
& 27B                    & 68.6$\pm$13.9 & 42.4$\pm$21.1 & 40.6$\pm$23.8 & 56.1$\pm$29.3 \\
& 35B-A3B                & 71.4$\pm$13.5 & \underline{47.4$\pm$20.9} & 40.3$\pm$20.3 & 64.9$\pm$25.9 \\
\cmidrule{1-6}
\multirow{1}{*}{GLM-4.6V-Flash}
& 9B                     & 55.7$\pm$6.4 & 29.6$\pm$17.1 & 20.4$\pm$13.7 & 74.9$\pm$23.1 \\
\cmidrule{1-6}
\multirow{5}{*}{InternVL3}
& 1B                     & 51.9$\pm$3.8 & 26.5$\pm$15.0 & 20.4$\pm$14.8 & 63.3$\pm$20.2 \\
& 2B                     & 53.3$\pm$5.6 & 25.4$\pm$16.0 & 20.4$\pm$14.2 & 50.9$\pm$23.6 \\
& 8B                     & 57.4$\pm$3.9 & 30.0$\pm$16.3 & 21.5$\pm$15.3 & 74.9$\pm$16.8 \\
& 14B                    & 55.1$\pm$4.6 & 28.8$\pm$16.6 & 21.3$\pm$15.0 & \textbf{76.0$\pm$21.1} \\
& 38B                    & 56.3$\pm$6.7 & 28.9$\pm$16.9 & 21.1$\pm$14.0 & \underline{75.1$\pm$21.6} \\
\cmidrule{1-6}
\multirow{7}{*}{InternVL3.5}
& 1B                     & 50.9$\pm$1.3 & 4.2$\pm$5.3 & 23.4$\pm$31.1 & 3.1$\pm$4.0 \\
& 2B                     & 53.6$\pm$4.3 & 25.6$\pm$15.7 & 21.2$\pm$16.0 & 43.9$\pm$14.2 \\
& 4B                     & 60.9$\pm$8.0 & 32.2$\pm$16.8 & 23.4$\pm$15.2 & 67.2$\pm$20.3 \\
& 8B                     & 60.1$\pm$10.5 & 28.9$\pm$17.7 & 29.8$\pm$22.0 & 46.8$\pm$26.4 \\
& 14B                    & 63.2$\pm$10.1 & 33.9$\pm$19.0 & 28.0$\pm$19.4 & 61.1$\pm$23.7 \\
& 30B-A3B                & 67.1$\pm$9.3 & 38.3$\pm$18.7 & 34.5$\pm$24.0 & 64.3$\pm$16.4 \\
& 38B                    & 64.9$\pm$11.2 & 36.5$\pm$17.6 & 29.0$\pm$20.7 & 73.3$\pm$15.9 \\
\cmidrule{1-6}
\multirow{2}{*}{Gemma-4}
& E2B                    & 50.6$\pm$1.6 & 6.3$\pm$12.0 & 12.1$\pm$17.2 & 5.8$\pm$12.0 \\
& 26B-A4B                & 67.9$\pm$13.9 & 38.3$\pm$23.6 & 39.8$\pm$25.5 & 55.6$\pm$30.8 \\
\midrule
RetGrounder & 8B & \underline{71.4$\pm$13.5} & 40.3$\pm$24.4 & 37.5$\pm$24.0 & 64.0$\pm$31.2 \\
\bottomrule
\end{tabular}%
}
\end{table}

\begin{figure}[tbp]
    \centering
    \includegraphics[width=\linewidth]{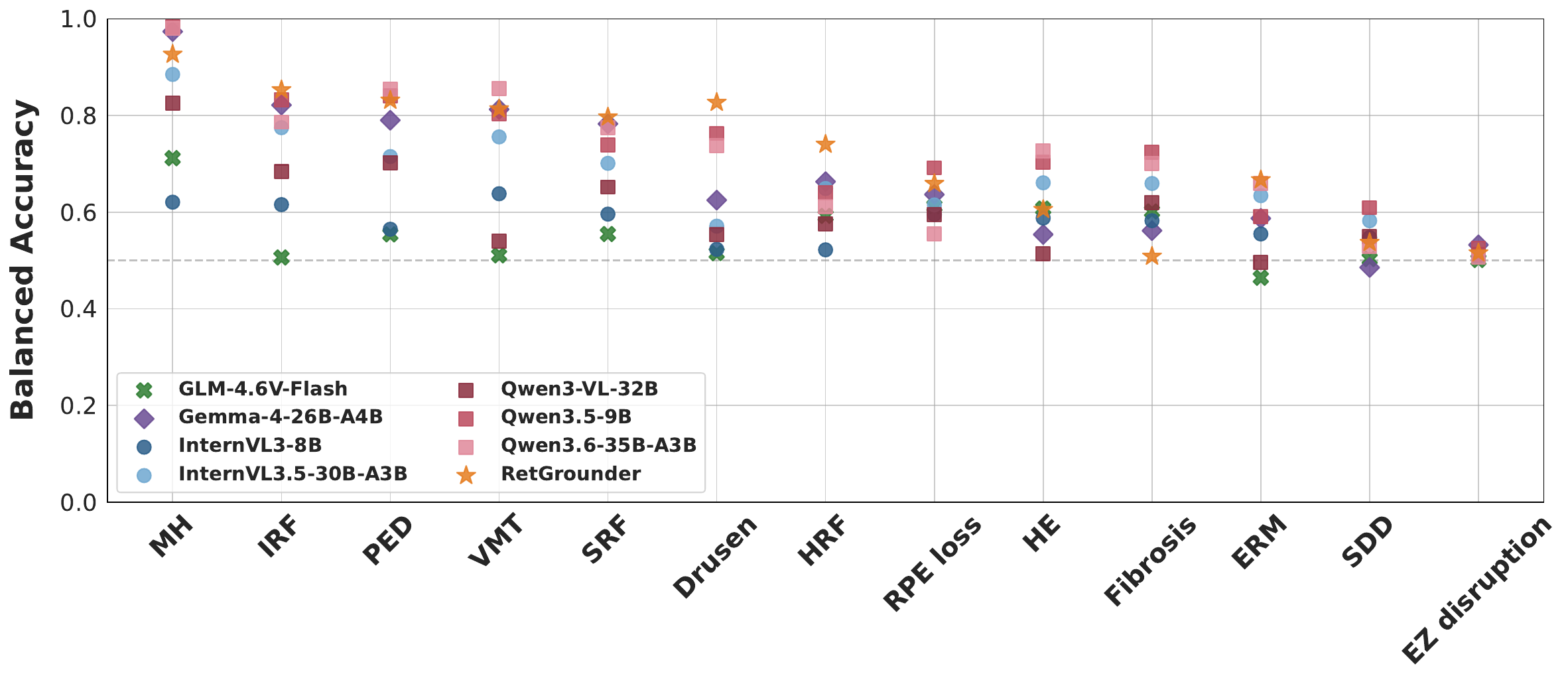}
    \caption{\textbf{Per-biomarker VQA balanced accuracy for the best-performing model in each family--generation.} Biomarkers are ordered by mean balanced accuracy (descending) across the displayed models.
    }
    \label{fig:vqa_biomarker_bacc}
\end{figure}

Per-biomarker VQA balanced accuracy (Fig.~\ref{fig:vqa_biomarker_bacc}) varied substantially across classes but followed a different pattern from grounding. Macular hole achieved the highest VQA performance (RetGrounder BAcc\,=\,92.7\%), consistent with its strong grounding results. However, drusen---among the lowest-performing classes for grounding (F1\,=\,28.7\%)---achieved high VQA balanced accuracy (82.7\%), indicating that models can recognise its presence without being able to delineate individual deposits. Conversely, fibrosis had the lowest VQA performance (BAcc\,=\,50.9\%, near chance) despite moderate grounding F1 (47.5\%). ERM (66.7\%) and hard exudates (60.5\%) occupied a middle tier with moderate performance.

\begin{figure}[tbp]
    \centering
    \includegraphics[width=\linewidth]{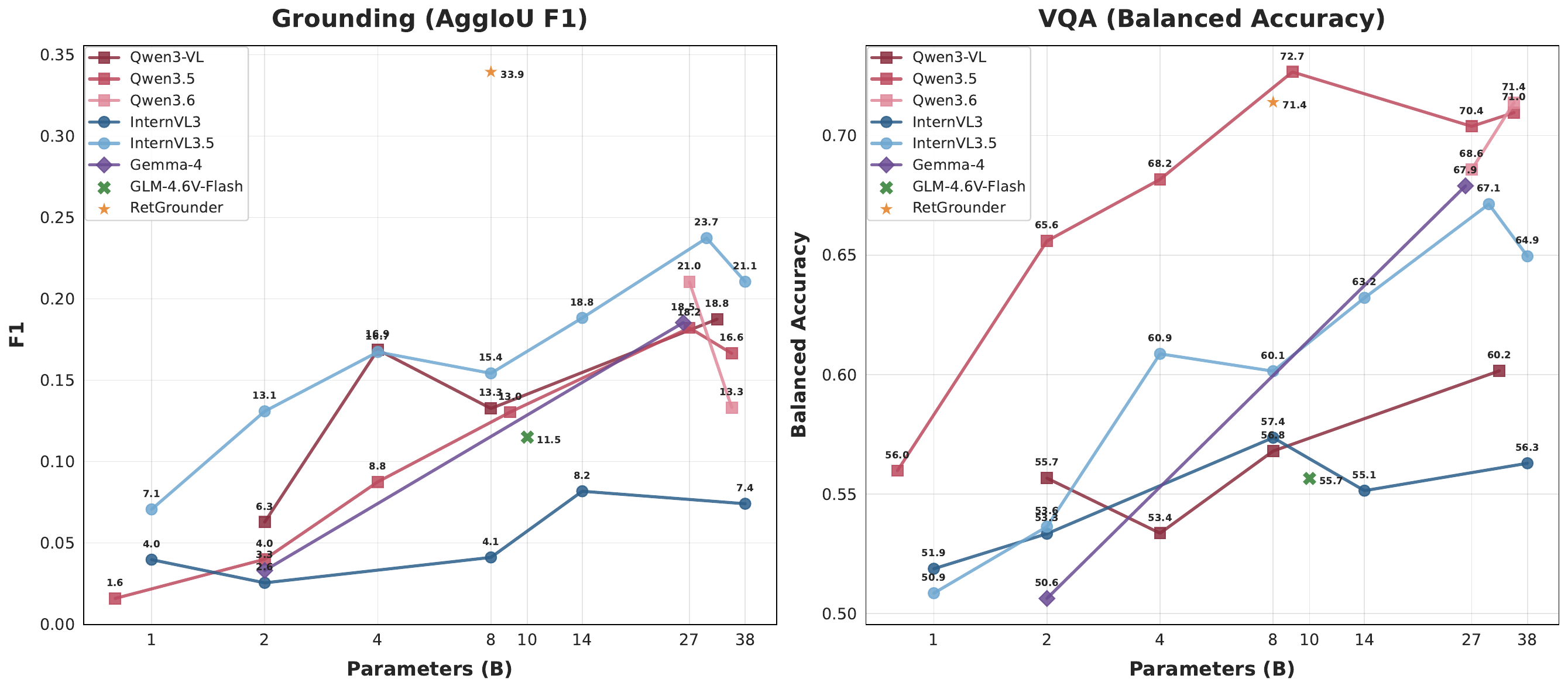}
    \caption{\textbf{Impact of model size and generation on grounding.} Performance is reported in terms of F1 (left) and VQA balanced accuracy (right).
    }
    \label{fig:scaling}
\end{figure}

\subsection{Grounding and VQA scale with model size, generation, and architecture}

Grounding performance generally improved with model scale in every family (Fig.~\ref{fig:scaling}), with larger models usually outperforming smaller models. However, the magnitude of the gains differed across model families and generations. Qwen3.5's performance rose, without a single reversal, from 0.8B (F1\,=\,1.6\%) to 27B (F1\,=\,18.2\%), and InternVL3.5 gained 16.6 F1 points between 1B (7.1\%) and 30B-A3B (23.7\%), whereas InternVL3 gained only 4.8 points across a nineteen-fold increase in parameters (2B, F1\,=\,2.6\%; 38B, F1\,=\,7.4\%). Nonetheless, there are exceptions to this general trend: Qwen3-VL-8B (F1\,=\,13.3\%) and InternVL3.5-8B (F1\,=\,15.4\%) both scored below their 4B counterparts (16.9\% and 16.7\%). Thus, while scale reliably predicts the direction of change within a family, it does not predict its magnitude nor the ordering of adjacent sizes. VQA balanced accuracy showed more predictable scaling, generally increasing with model size across all families, with the best models exceeding 70\%.

Improvements over model generations were overall more consistent. Qwen3.6-27B (F1\,=\,21.0\%) outperformed Qwen3.5-27B (F1\,=\,18.2\%) and Qwen3-VL-32B (F1\,=\,18.8\%). InternVL3.5 models outperformed their InternVL3 counterparts at every size, with especially large gains at smaller scales (InternVL3.5-2B F1\,=\,13.1\% vs.\ InternVL3-2B F1\,=\,2.6\%). However, our fine-tuned 8B RetGrounder still outperformed all open-source models up to 38B parameters, highlighting the benefits of specialization.

Three models included both a dense and a mixture-of-experts (MoE) checkpoint of comparable total size: Qwen3.5 (27B dense vs.\ 35B-A3B MoE), Qwen3.6 (27B dense vs.\ 35B-A3B MoE), and InternVL3.5 (38B dense vs.\ 30B-A3B MoE). The MoE checkpoint improved VQA balanced accuracy over its dense counterpart in all three cases (Qwen3.5: 71.0\% vs.\ 70.4\%; Qwen3.6: 71.4\% vs.\ 68.6\%; InternVL3.5: 67.1\% vs.\ 64.9\%). Its effect on grounding F1 was less consistent, improving over the dense checkpoint for InternVL3.5 (23.7\% vs.\ 21.1\%) but not for Qwen3.5 (16.6\% vs.\ 18.2\%) or Qwen3.6 (13.3\% vs.\ 21.0\%).

\subsection{Effect of training data composition}

\begin{figure}[tbp]
    \centering
    \includegraphics[width=0.6\linewidth]{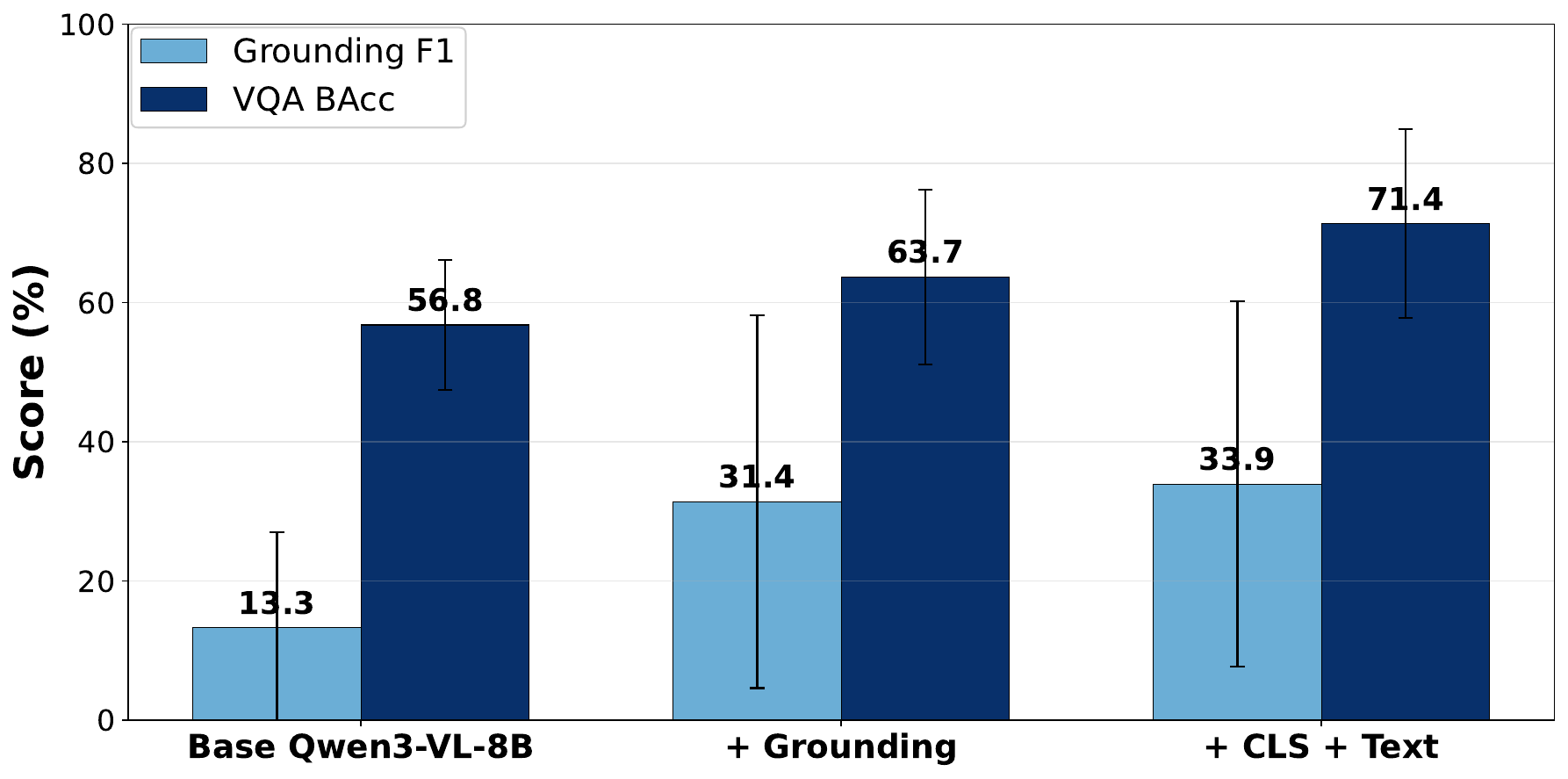}
    \caption{\textbf{Ablation study of RetGrounder's training data.}
    QA denotes VQA classification data and text-only QA.
    Base is the unmodified Qwen3-VL-8B-Instruct.}
    \label{fig:ablation}
\end{figure}

RetGrounder is fine-tuned on two kinds of retinal OCT data: bounding-box \emph{grounding} data, and
question--answering (QA) supervision, comprising classification data recast into VQA form.
It is also trained on text-only QA pairs covering biomarker knowledge (see Methods for more details). To establish what each contributes, we fine-tuned the base Qwen3-VL-8B-Instruct on the grounding component alone and on the full mixture, and evaluated both against the unmodified base model. Fig.~\ref{fig:ablation} reports grounding F1 and VQA balanced accuracy for the three resulting models.

Both components contributed, and on both tasks. Fine-tuning on grounding data alone more than doubled grounding F1 relative to the base model (13.3\%\,$\rightarrow$\,31.4\%) and also lifted VQA balanced accuracy (56.8\%\,$\rightarrow$\,63.7\%), indicating that learning to localise biomarkers improves presence detection as well. Adding the
QA
component raised both further, to F1\,=\,33.9\% and BAcc\,=\,71.4\%. The two components divide the labour unevenly across the two tasks: grounding data supplies most of the 20.6-point grounding gain (18.1 points, against 2.5 from
QA), whereas the 14.6-point VQA gain is split almost equally between them (6.9 and 7.7 points). Grounding supervision is thus what makes the model localise, but it is not the only source of its VQA ability.

\subsection{Inter-observer agreement}

To establish a human reference for grounding performance and to test the benchmark's stability to annotation variability, a subset of 84 B-scans (63 from OCTDL, 21 from OLIVES; at least 20 images per biomarker) was independently re-annotated in a second reading following the same annotation protocol. We treated the two readings as two observers and scored the second read against the original with the same AggIoU matching used for the models (Methods). 

Inter-observer agreement reached a macro F1 of 41.0\% (39.9\% averaged over both comparison directions) and a mean AggIoU of 26.0\% (per-class values in Supplementary Tables~\ref{tab:perclass_aggiou_84} and~\ref{tab:perclass_f1_84}), exceeding the grounding performance of the best model on the same images (RetGrounder F1\,=\,31.2\%; 33.9\% on the full benchmark). RetGrounder therefore approaches, but does not reach, the level of agreement between human annotators. Agreement was high for well-circumscribed lesions (macular hole F1\,=\,90.0\%, vitreomacular traction 77.3\%, PED 72.5\%, SRF 70.6\%) and low for epiretinal membrane (15.7\%). For epiretinal membrane the disagreement was one of presence rather than localisation: the two readings differed in how often the biomarker was marked at all (presence-level Cohen's~$\kappa$\,=\,0.37, against a mean of 0.55 across the thirteen classes).

The benchmark's model ranking was insensitive to which reading was used as reference: scoring all models against the second reading rather than the original left their ordering essentially unchanged (Spearman~$\rho$\,=\,0.956 across the 28 models, with RetGrounder ranked first under both) and shifted scores downward by a similar margin for every model (mean 2.2 F1 points). The conclusions drawn from RetinaGroundBench therefore do not depend on the annotations of a single reading.

\section{Discussion}

In this work, we introduced RetinaGroundBench, a benchmark of 815 retinal OCT B-scans annotated with 4{,}537 bounding boxes across 13 biomarker classes, and used it to systematically evaluate the visual grounding and VQA capabilities of 27 state-of-the-art open-source MLLMs, spanning the most capable model families available at the time of writing. We also proposed AggIoU, both an overlap metric and a matching criterion tailored to the multi-box prediction behaviour of MLLMs, and presented RetGrounder, a domain-adapted 8B model fine-tuned on retinal OCT data.

Our results reveal that visual grounding of retinal OCT biomarkers remains a challenging and largely unsolved task for general-purpose open-source MLLMs. Even the best general-purpose open-source model achieved a macro F1 of only 23.7\%, and all models were characterised by low recall, indicating that they do not yet localise the majority of biomarker instances. The conservative prediction behaviour observed in models such as Gemma-4-26B-A4B---high precision at the cost of very low recall---suggests that these models default to producing few, safe predictions rather than attempting comprehensive localisation, even though the grounding prompt explicitly instructs them to locate every instance of the queried biomarker (Supplementary Table~\ref{tab:prompts}).
The substantial heterogeneity in per-biomarker performance likely reflects differences in visual distinctiveness across pathology types. Well-circumscribed pathologies such as macular hole present an unambiguous visual signature, i.e., a full-thickness disruption of the retinal layers that is recognisable even without domain-specific training. In contrast, subtle, diffuse or texture-based biomarkers such as hard exudates and drusen present more irregular or poorly defined spatial extents, with higher visual similarity to normal tissue.

The stark contrast between grounding and VQA performance indicates that these two capabilities are not tightly coupled in current MLLMs. RetGrounder dominates grounding yet ranks behind the leading general-purpose models on VQA; conversely, the Qwen3.5 and Qwen3.6 models lead VQA without achieving proportionally strong grounding. The VQA precision--recall landscape shows how some of that VQA strength arises: several models achieved high recall by predicting biomarker presence almost indiscriminately, yielding low precision and inflating balanced accuracy---a metric that weights sensitivity and specificity equally---in a way that would be problematic in deployment, where false positives carry costs. Strong presence detection therefore does not entail strong localisation, nor the reverse; image-level pattern recognition and spatial localisation appear to be distinct capabilities requiring different forms of supervision. Clinically, this means both axes must be evaluated explicitly rather than either being taken as a proxy for the other.

The same dissociation appears at the class level, where correspondence between the two tasks is only partial. Macular hole, the highest-scoring class for grounding (F1\,=\,95.1\%), also scored highest for VQA (BAcc\,=\,92.7\%), consistent with its unambiguous visual signature. However, the correspondence breaks down elsewhere: drusen reached 82.7\% BAcc despite being among the hardest classes to ground (F1\,=\,28.7\%), indicating that models can recognise drusen from global image features without delineating individual deposits, while fibrosis sat effectively at chance for VQA (BAcc\,=\,50.9\%) yet achieved moderate grounding F1 (47.5\%), suggesting that when models detect fibrosis at all they can localise it, but that the class is frequently missed entirely. Several factors may underlie this dissociation. Localisation is inherently more demanding than binary presence detection, requiring the model to resolve fine spatial extents rather than aggregate global features. The text-based coordinate format used by current MLLMs may itself be a bottleneck: the vision encoder may capture spatial information that is then lost or degraded when serialised as coordinate tokens through an autoregressive language decoder. Whether the grounding--VQA gap reflects a representational limitation, an architectural mismatch between task and output format, or simply the greater intrinsic difficulty of localisation remains an open question.

Regarding architecture, MoE checkpoints offered stronger VQA performance than their dense counterparts within each model generation, though the effect on grounding was inconsistent.
Further architectural explanations could be attempted by
comparing models from different families; however,
the differences in both architecture and training data of open-source models make
any such account associative at best. We therefore restrict ourselves to observations.
Strong grounding did not require a large vision encoder: InternVL3.5-30B-A3B, the best general-purpose grounding model in our evaluation, uses one of only 300M parameters. Conversely, the newer Qwen3.5, which introduced hybrid Gated DeltaNet attention, improved VQA markedly while regressing grounding at smaller scales relative to Qwen3-VL; advances in general vision--language capability evidently do not transfer automatically to spatial localisation. What the comparison does establish unambiguously is that none of these differences closes the gap to clinical usefulness: RetGrounder's 8B model outperformed every open-source model up to 38B parameters on grounding. Thus, targeted fine-tuning on retinal data compensated for a nearly fivefold difference in capacity.

The ablation indicates that the two training components during RetGrounder's fine-tuning play asymmetric roles. Grounding supervision accounts for the bulk of the localisation gain (18.1 of 20.6 F1 points), whereas the VQA gain divides almost equally between the two. Notably, each component also improves the task it is not aimed at: grounding data alone raises VQA balanced accuracy by nearly seven points, suggesting that learning where a biomarker lies reinforces judging whether it is present, and
QA in turn contributes to localisation. The two forms of supervision are therefore complementary rather than redundant. For domain adaptation this argues against treating spatial annotation as optional: it carries most of the localisation signal in our mixture, and image-level labels and text knowledge, though cheaper to obtain, close only a small part of that gap. 

While RetinaGroundBench provides a reproducible reference point for grounded biomarker evaluation in retinal OCT, with RetGrounder narrowing the gap to clinically useful localisation, several aspects remain open.
We set them out below as areas for improvement, each with a concrete next step. 

In terms of performance,
RetGrounder has shown suboptimal results for two lesions, fibrosis and RPE loss.
The former is likely a cost of specialisation: fibrosis has no close analogue among the classes RetGrounder was fine-tuned on (Supplementary Table~\ref{tab:internal_grounding}), so the general-purpose InternVL3.5-30B-A3B, with no such prior, localises it marginally better (F1\,=\,49.4\% vs.\ 47.5\%).
The low performance for RPE loss, on the other hand, is generalized, with near-floor F1 scores for every model.
Adapting a model to a lesion inventory therefore improves accuracy on those lesions at some cost on unfamiliar targets, which argues for curating fine-tuning data to span a broader range of lesions rather than for abandoning domain adaptation.

Regarding fine-tuning data, two main limitations exist. First, the benchmark comprises 815 B-scans drawn from two public datasets (OCTDL and OLIVES) acquired on a limited range of imaging devices, and each image is a single 2D B-scan rather than a full volumetric scan; it may therefore not capture the full diversity of scanner hardware, acquisition protocols, and patient demographics encountered in clinical practice. Expanding the benchmark across additional datasets, imaging devices, and volumetric context would strengthen its representativeness and support more fine-grained analyses.
Second, the bounding-box paradigm itself presents some limitations for the representation of lesions. For instance, drusen and hard exudates commonly present as multiple small lesions scattered across the retina, and there is no canonical way to group such instances into one box versus several. Annotators are known to differ in this granularity~\cite{lepetit2026retinal}, which likely contributes to the low inter-observer agreement and the low AggIoU observed for some classes, placing a practical ceiling on achievable overlap.
This issue could be mitigated by the use of pixel-level segmentation masks, which are more standardized and represent a more natural fit.
Moreover, evaluating with this type of annotation would establish how much of the observed shortfall is attributable to the annotation format rather than to the models themselves.

Finally, the evaluation also presents some limitations. First, we evaluate only open-source models with native bounding-box output. Frontier closed-source systems are excluded for two reasons. Most do not natively emit bounding-box coordinates, and so cannot be assessed on grounding.
Furthermore, none can be used locally, but only via APIs over internet connections, limiting their applicability to clinical setups, where data privacy is a key concern.
Relatedly, closed-source endpoints are subject to unannounced updates, deprecation, and removal, so results obtained from them cannot be version-controlled or reliably reproduced, hindering benchmarking.
And second, because the evaluated model families differ simultaneously in architecture and training data,
it is not possible to analyse, in a disentangled manner, the impact of each component on model performance.
For this, further controlled studies that vary architecture while holding training data fixed would be needed.

Taken together, our results lead to three main findings. First, strong VQA performance does not imply strong grounding, and vice versa. Models that reliably detect biomarker presence may fail to localise them, and the reverse also occurs.
This implies that clinical deployment requires explicit evaluation of both axes. Second, model scale alone is insufficient for medical visual grounding.
Increasing parameters generally improved grounding across different families, but by highly variable margins,
and the best general-purpose model at 30B remained far below clinical usability.
Third, targeted domain adaptation remains the most effective path to clinically useful grounding: RetGrounder demonstrates that a modestly sized model trained on curated retinal data can substantially outperform much larger general-purpose models. This, however, represents a first step rather than a final solution, and room for improvement remains before clinical applicability; larger-scale curated retinal pretraining, more sophisticated data-mixing curricula, and ultimately prospective clinical reader studies are the natural next steps.

To conclude, for an MLLM to be trusted in the clinic, it is not enough that its answers are correct; a clinician must be able to see the evidence behind them. This work provides the most comprehensive assessment of whether current MLLMs meet that standard on retinal OCT, and contributes the three components needed to ask the question: RetinaGroundBench, a benchmark pairing expert biomarker annotations with a text-queried grounding protocol; AggIoU, a metric that scores localisation as MLLMs actually perform it, by aggregating the multiple boxes they emit rather than penalising them; and RetGrounder, which provides a straightforward recipe for narrowing that gap.
The answer, for now, is that state-of-the-art open-source models largely cannot ground their answers on OCT---but that targeted domain adaptation closes much of the distance at a fraction of the parameter count. By releasing the benchmark, the annotations, and the metric, we provide a fixed reference point against which the field can measure that progress as models advance.

\section{Methods}

\subsection{RetinaGroundBench}

We evaluate open-source MLLMs across two axes: visual grounding and visual question answering. To do this, we propose RetinaGroundBench, a dataset composed of 815 retinal OCT images from the public OCTDL~\cite{Kulyabin2024} and OLIVES~\cite{prabhushankarolives2022} datasets, annotated with bounding boxes for 13 biomarker classes: pigment epithelial detachment (PED), intraretinal fluid (IRF), subretinal fluid (SRF), drusen, epiretinal membrane (ERM), fibrosis, hard exudates, macular hole, vitreomacular traction, ellipsoid zone disruption, subretinal drusenoid deposits, hyperreflective foci, and retinal pigment epithelium (RPE) loss. In total, the benchmark contains 4{,}537 ground truth bounding boxes. Table~\ref{tab:gt_distribution} reports the per-class distribution of annotated images and boxes, together with the mean box area of each biomarker.

\begin{table}[tbph]
\centering
\caption{\textbf{Ground truth biomarker distribution across the 815 retinal OCT images of RetinaGroundBench.} Count indicates the number of images containing each biomarker; Boxes the total number of GT bounding boxes; Avg.\ Area the mean bounding box area in pixels. All 13 biomarker classes are included in the grounding and VQA evaluation.
In total, the benchmark consists of 4{,}537 bounding boxes.
}
\label{tab:gt_distribution}
\begin{tabular}{lrrrr}
\toprule
\textbf{Biomarker} & \textbf{Images} & \textbf{\%} & \textbf{Boxes} & \textbf{Avg.\ Area (px)} \\
\midrule
Intraretinal fluid             & 391 & 48.0 & 1,045 &  6,723 \\
Hyperreflective foci           & 302 & 37.1 & 1,114 &    987 \\
Subretinal fluid               & 200 & 24.5 &   347 &  6,119 \\
Hard exudates                  & 184 & 22.6 &   391 &  1,030 \\
Pigment epithelial detachment  & 174 & 21.3 &   330 & 16,191 \\
Ellipsoid zone disruption      & 149 & 18.3 &   229 &  2,027 \\
Drusen                         & 136 & 16.7 &   315 &  1,996 \\
Fibrosis                       & 132 & 16.2 &   187 & 17,242 \\
Epiretinal membrane            & 121 & 14.8 &   375 &  3,635 \\
Subretinal drusenoid deposits  &  55 &  6.7 &   100 &  1,738 \\
Macular hole                   &  32 &  3.9 &    32 & 11,748 \\
Retinal pigment epithelium loss &  29 &  3.6 &    45 &  5,508 \\
Vitreomacular traction         &  21 &  2.6 &    27 & 12,050 \\
\bottomrule
\end{tabular}
\end{table}

\begin{figure}[tbp]
    \centering
    \includegraphics[width=\linewidth,height=0.85\textheight,keepaspectratio]{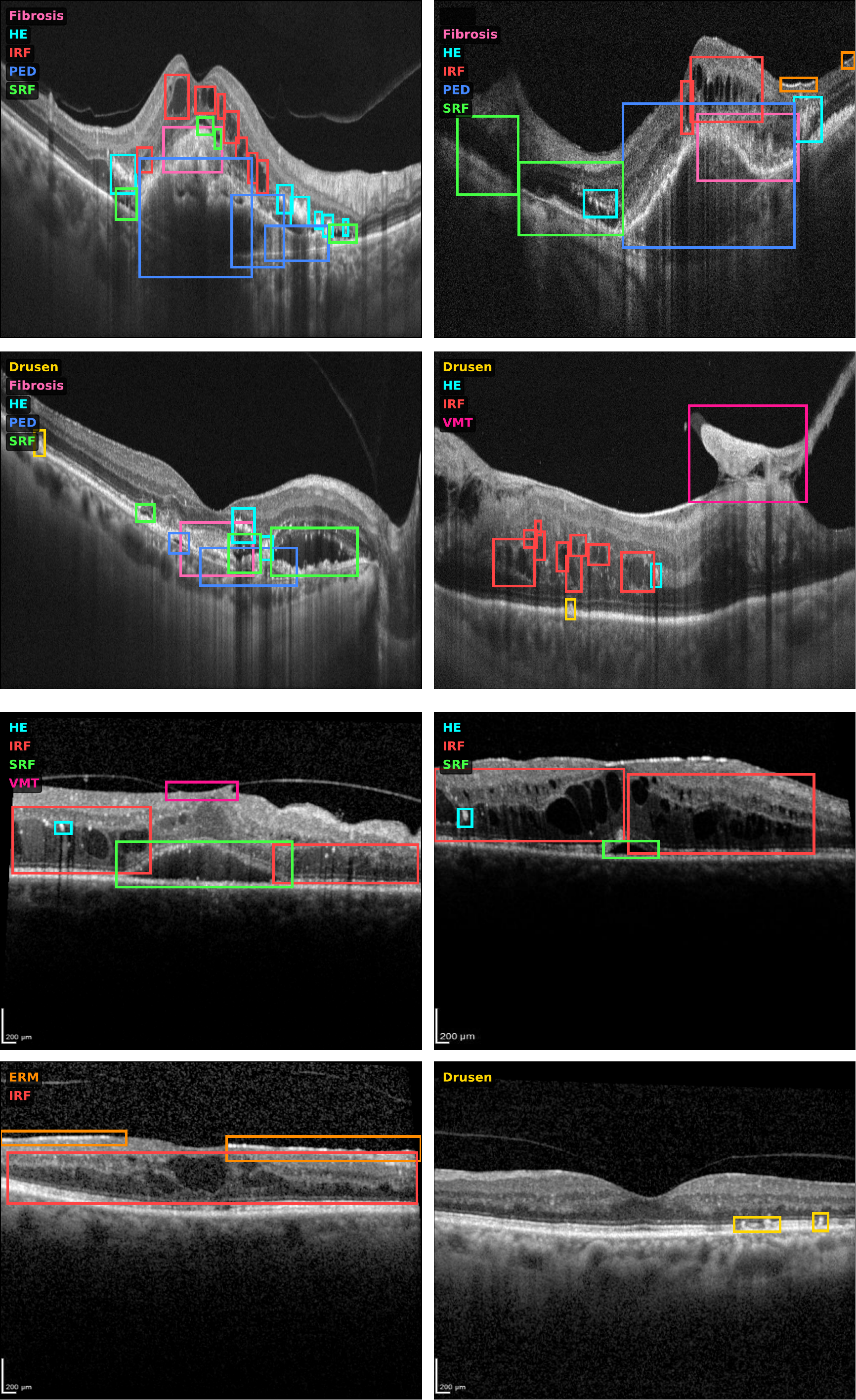}
    \caption{\textbf{Example annotated images from the RetinaGroundBench benchmark.} Rows 1--2: OCTDL images; rows 3--4: OLIVES images. Bounding boxes are colour-coded by biomarker class.}
    \label{fig:annotation_examples}
\end{figure}

Annotations were produced by a team of three ophthalmologists from the Medical University of Vienna. Two junior ophthalmologists independently labelled separate subsets of images, each drawing bounding boxes around all visible instances of the 13 biomarker classes. A senior retina specialist then reviewed, corrected, and approved all annotations. This two-stage workflow---independent junior labelling followed by senior expert adjudication---ensured both annotation throughput and clinical accuracy. Representative annotated images from both source datasets are shown in Fig.~\ref{fig:annotation_examples}.

Evaluated models were selected based on three criteria: (i)~availability of open weights, (ii)~explicit support for pixel-level visual grounding with bounding box coordinate output, and (iii)~a parameter count of at most 38B. The last criterion reflects the constraint that patient data generally cannot leave the institution, so a model intended for clinical use must be self-hosted on hardware a hospital can realistically operate; we therefore excluded models too large to serve on a single GPU. We restrict the evaluation to open-weight models for the same data privacy concerns but also reproducibility: closed-source systems are subject to unannounced updates, deprecation, and removal, whereas open weights are permanently available and version-controlled, ensuring the benchmark results remain replicable.

Applying these criteria yields 27 open-source models released up to 30 April 2026, spanning four model families and multiple generations---Qwen3-VL (2B--32B), Qwen3.5 (0.8B--35B), Qwen3.6 (27B--35B), InternVL3 (1B--38B), InternVL3.5 (1B--38B), GLM-4.6V-Flash (9B), and Gemma-4 (2B--26B). We additionally evaluate RetGrounder, a Qwen3-based 8B model we fine-tuned on retinal imaging data. Supplementary Table~\ref{tab:model_repos} lists the exact repository for each evaluated model.

\subsection{Aggregated IoU (AggIoU)}

\begin{figure}[tbp]
\centering
\includegraphics[width=\linewidth]{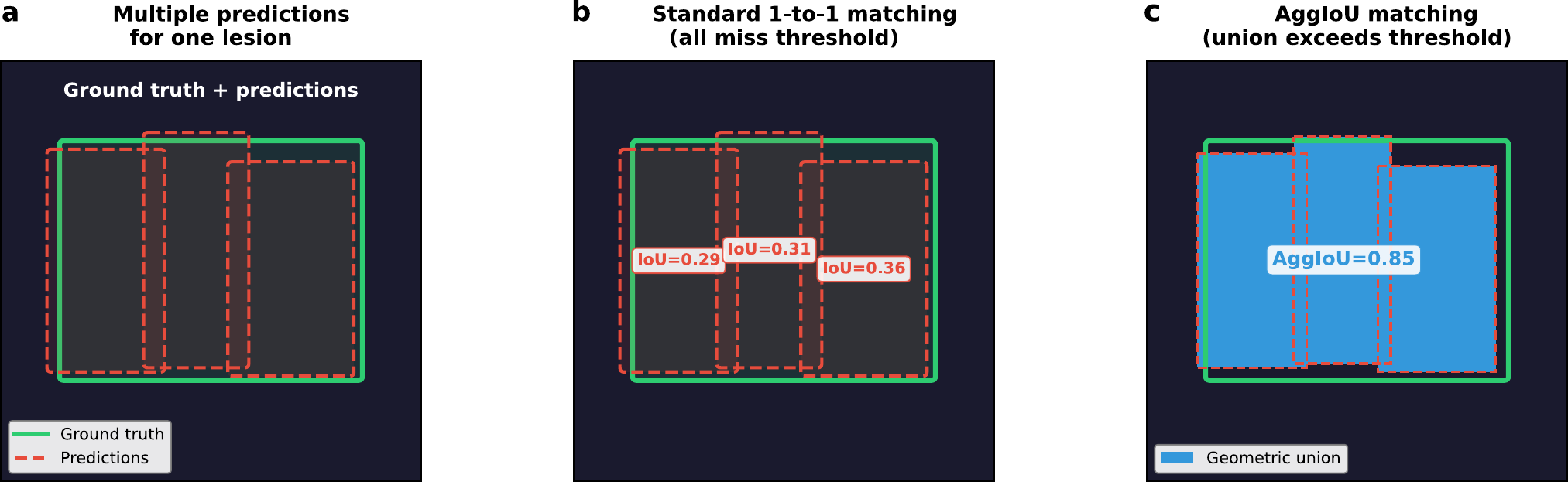}
\caption{\textbf{Illustration of the AggIoU matching strategy.}
\textbf{a.}~An MLLM produces multiple predicted boxes (red, dashed) for a single ground truth region (green).
\textbf{b.}~Standard one-to-one matching evaluates each prediction independently; all fall below the IoU threshold and the lesion is counted as a false negative. 
Precision/Recall = 0.33/1.0.
\textbf{c.}~AggIoU computes the geometric union of gathered predictions, yielding a high overlap that correctly identifies the lesion as a true positive.
Precision/Recall = 1.0/1.0.
}
\label{fig:aggiou_schematic}
\end{figure}

Standard object detection metrics such as mean average precision rely on one-to-one matching between predictions and ground truth boxes, penalising models that produce multiple overlapping predictions for a single lesion. In retinal OCT imaging, MLLMs frequently predict several small boxes that collectively cover a ground truth region (Fig.~\ref{fig:aggiou_schematic}a). Under standard matching, each prediction is evaluated independently and may fall below the IoU threshold despite the set of predictions adequately covering the lesion (Fig.~\ref{fig:aggiou_schematic}b). To accommodate this behaviour, we propose Aggregated IoU (AggIoU), which aggregates predictions per ground truth box and serves both as an overlap metric and as a matching criterion (Fig.~\ref{fig:aggiou_schematic}c).

The matching proceeds as follows for each image and biomarker class; Algorithm~\ref{alg:aggiou} states the same procedure in full, including the macro-averaging described below. For each ground truth box, we gather all predicted boxes whose intersection with the ground truth exceeds a fraction $t_1$ of the minimum of their areas:
\begin{equation}
    \text{area}(\text{GT} \cap \text{pred}) > t_1 \cdot \min(\text{area}(\text{GT}),\, \text{area}(\text{pred})).
\end{equation}
The geometric union of the gathered predictions is then computed exactly, and we define the \emph{AggIoU} score of the ground truth box as the intersection-over-union (IoU) between it and this union. If the AggIoU score exceeds a second threshold $t_2$, the ground truth is counted as a true positive; otherwise it is a false negative. Any predicted box not gathered by any ground truth is counted as a false positive.

\begin{algorithm}[ht]
\caption{AggIoU matching and macro-averaged metric computation.}\label{alg:aggiou}
\begin{algorithmic}[1]
\REQUIRE ground-truth boxes $\mathcal{G}$ and predicted boxes $\mathcal{P}$ (per image and class); thresholds $t_1,t_2$; number of classes $C$
\ENSURE macro-averaged precision $\bar{P}$, recall $\bar{R}$, and F1 $\overline{\mathrm{F1}}$
\STATE $\mathrm{TP}_c,\mathrm{FP}_c,\mathrm{FN}_c \gets 0$ for all $c \in \{1,\dots,C\}$
\FOR{each image and each class $c$}
    \STATE $\mathrm{matched} \gets \emptyset$
    \FOR{each ground-truth box $g \in \mathcal{G}_c$}
        \STATE $\mathrm{gathered} \gets \emptyset$
        \FOR{each predicted box $p \in \mathcal{P}_c$}
            \IF{$\mathrm{area}(g \cap p) > t_1 \cdot \min(\mathrm{area}(g),\, \mathrm{area}(p))$}
                \STATE $\mathrm{gathered} \gets \mathrm{gathered} \cup \{p\}$, \quad $\mathrm{matched} \gets \mathrm{matched} \cup \{p\}$
            \ENDIF
        \ENDFOR
        \STATE $U \gets \textsc{GeometricUnion}(\mathrm{gathered})$
        \STATE $\mathrm{AggIoU}(g) \gets \mathrm{IoU}(g, U)$
        \IF{$\mathrm{AggIoU}(g) \ge t_2$}
            \STATE $\mathrm{TP}_c \gets \mathrm{TP}_c + 1$
        \ELSE
            \STATE $\mathrm{FN}_c \gets \mathrm{FN}_c + 1$
        \ENDIF
    \ENDFOR
    \STATE $\mathrm{FP}_c \gets \mathrm{FP}_c + |\mathcal{P}_c \setminus \mathrm{matched}|$
\ENDFOR
\FOR{$c = 1$ to $C$}
    \STATE $P_c \gets \mathrm{TP}_c / (\mathrm{TP}_c + \mathrm{FP}_c)$
    \STATE $R_c \gets \mathrm{TP}_c / (\mathrm{TP}_c + \mathrm{FN}_c)$
    \STATE $\mathrm{F1}_c \gets 2\,P_c R_c / (P_c + R_c)$
\ENDFOR
\STATE \textbf{return} $\bar{P} = \tfrac{1}{C}\sum_c P_c$,\ \ $\bar{R} = \tfrac{1}{C}\sum_c R_c$,\ \ $\overline{\mathrm{F1}} = \tfrac{1}{C}\sum_c \mathrm{F1}_c$
\end{algorithmic}
\end{algorithm}

\subsection{Evaluation metrics}

Per-class precision, recall, and F1 are computed from true positive, false positive, and false negative counts accumulated over all images in the combined OCTDL and OLIVES evaluation set; the two source datasets are pooled rather than scored separately. In addition, we report the mean AggIoU score---the per-box ground-truth--union overlap averaged over all ground-truth boxes---as a direct measure of spatial localisation quality independent of the $t_2$ threshold. All reported grounding metrics are macro-averaged: per-class scores are computed first, then averaged across all 13 biomarker classes to ensure equal contribution from each class regardless of prevalence.
The two thresholds are set once and held fixed for every model; the values used, and their effect on the results, are given under Implementation details.

For visual question answering, we prompt each model with each of the 13 biomarker classes for every image in the benchmark (815 images), asking whether the given class is present. We compute the balanced accuracy for each class as the mean of sensitivity and specificity, then macro-average across all 13 classes to produce the final metric. This ensures equal contribution from each biomarker, consistent with the grounding evaluation.

\subsection{RetGrounder}

For developing RetGrounder, we fine-tune Qwen3-VL-8B-Instruct with LoRA~\cite{hu2022lora}, which restricts training to low-rank adapters in order to save computation and avoid catastrophic forgetting of the base model's general capabilities. Training uses a mixture of datasets covering three task types: visual grounding, VQA-style classification, and text-only question-answer pairs. Supplementary Table~\ref{tab:training_data} summarises the training data composition, and training hyperparameters are given in Implementation details below.

\paragraph{In-house grounding dataset.}
The primary grounding data source is an in-house dataset of 34{,}112 OCT B-scans extracted from 934 volumetric OCT scans acquired on Spectralis devices (Heidelberg Engineering). The volumes span geographic atrophy, intermediate AMD, and neovascular AMD cohorts. Pixel-level segmentation masks for 8 lesion classes (reticular pseudodrusen, IRF, hyperreflective foci, retinal pigment epithelium elevations, SRF, subretinal hyperreflective material, outer retinal tubulation, and hypertransmission) were produced by trained graders using a semi-automated pipeline: an initial segmentation model generated candidate masks, which were then manually reviewed and corrected by expert graders. Not all classes are annotated in every image.
To convert pixel-level masks to bounding boxes, we applied a contour-detection and clustering approach. For each class, connected components were extracted from the binary mask, discarding regions below a minimum area threshold; spatially proximate contours of the same class were then merged, to avoid fragmenting a single lesion instance into multiple small boxes, and the enclosing bounding box of each cluster was taken as the final annotation. All bounding box coordinates were normalised to a $[0, 1000]$ coordinate system. Supplementary Table~\ref{tab:internal_grounding} details the per-class distribution, totalling 154{,}548 bounding boxes.

\paragraph{Public grounding datasets.}
We also convert several public segmentation datasets to bounding box labels using the same rule-based approach. We use 1{,}199 images from OIMHS~\cite{Ye2023} covering IRF and macular hole, 3{,}012 images from RVO-ME~\cite{Xiong2026} covering IRF, SRF, external limiting membrane, ellipsoid zone, and hyperreflective foci, and 928 images from OCTAVE~\cite{octave} covering ERM. We also use 566 images (4{,}698 bounding boxes) from OCT5k~\cite{Arikan2025} which already come in bounding box form, covering hard drusen, soft drusen, drusenoid PED, photoreceptor layer disruption, reticular pseudodrusen, choroidal folds, geographic atrophy, hyper fluorescent spots, and IRF.

\paragraph{Classification datasets.}
Classification datasets are converted into VQA format where the model is prompted to identify the disease or lesion present in the image. We use all 24{,}000 images from C8~\cite{obuli_sai_naren_2021} covering AMD, CNV, CSR, DME, DR, drusen, macular hole, and normal retina; 16{,}979 images from Kermany~\cite{Kermany2018} covering CNV, DME, drusen, and normal cases; all 572 samples from OCTID~\cite{GHOLAMI2020106532} covering AMD, CSR, DR, macular hole, and healthy retinas; and 4{,}205 images from an in-house classification set covering RVO, CNV, Stargardt disease, iAMD, GA, DME, and healthy retinas.

\paragraph{Text-only QA.}
We generate 300 text-only question-answer pairs covering OCT biomarker knowledge without any accompanying images. Generation proceeded iteratively using three frontier LLMs (Claude, ChatGPT, and Gemini): each model was prompted to produce question-answer pairs spanning biomarker definitions, inter-class relationships, differential diagnosis, spatial localisation on OCT B-scans, and clinical significance. The 230 general OCT knowledge pairs cover topics such as the relationship between related biomarker classes (e.g.\ PED vs.\ RPE elevations), OCT appearance descriptions, and differential features. An additional 70 spatial reasoning pairs test anatomical layer relationships (e.g.\ ``If intraretinal fluid is in the outer plexiform layer and subretinal fluid is below the photoreceptors, which is deeper?''). After generation, each pair was cross-checked by the other two LLMs for factual accuracy; pairs flagged as incorrect by any model were revised or discarded.
Supplementary Table~\ref{tab:prompts} shows representative examples of the prompts used for evaluation and training across the different task types.

\subsection{Implementation details}
RetGrounder was trained with the LlamaFactory~\cite{zheng2024llamafactory} library for 77k steps with an effective batch size of 8, an initial learning rate of $10^{-5}$ with a 0.1 warm-up ratio, and a cosine-with-restart learning rate scheduler. LoRA was applied to the entire model with a rank of 64 and an alpha of 128, using a 90--10\% train--validation split.

Mask-to-bounding-box conversion for the in-house and public segmentation datasets used OpenCV contour detection to extract connected components, discarding regions below a minimum area threshold, followed by DBSCAN clustering with a distance threshold of 25 pixels to merge spatially proximate contours of the same class.

AggIoU was computed with $t_1 = 0.1$ and $t_2 = 0.3$ for all models: a permissive gathering threshold allows even small overlapping fragments to contribute to the union, while the moderate acceptance threshold ensures that the aggregated prediction still covers a meaningful portion of the ground truth. An internal sweep over threshold pairs confirmed that the ranking of models is stable across a wide range of settings.

Inference for every evaluated model family was run from the model's public Hugging Face repository (Supplementary Table~\ref{tab:model_repos}) using each family's reference implementation. A self-contained pipeline reproducing the full grounding evaluation is available at the repository given under Code availability.

\section{Author contributions}
R.F. conceived the study, designed the benchmark, implemented the evaluation pipeline and the RetGrounder model, performed the experiments and analysis, and wrote the manuscript. J.M. contributed to the methodology, AggIoU metric design, and critically revised the manuscript. B.F. contributed to training data acquisition and preparation as well as metric design. P.W.-S. and D.H. provided clinical expertise and performed the retinal OCT biomarker annotations. L.H. and V.M. contributed to data curation and annotation review. A.P. provided ophthalmological guidance and clinical validation of the annotations. H.B. supervised the project, contributed to its conception and methodology, acquired funding, and critically revised the manuscript. All authors reviewed and approved the final manuscript.

\section{Data availability}

The RetinaGroundBench benchmark is derived from two publicly available datasets: OCTDL (\url{https://github.com/MikhailKulyabin/OCTDL}) and OLIVES (\url{https://doi.org/10.5281/zenodo.7105232}). Our bounding box annotations and evaluation code are publicly available at \url{https://github.com/ronnief1/RetinaGroundBench}.

The RetGrounder training mixture additionally draws on seven public datasets: OIMHS (\url{https://doi.org/10.6084/m9.figshare.23508453.v1}), RVO-ME (\url{https://doi.org/10.6084/m9.figshare.29804435.v1}), OCTAVE (\url{https://doi.org/10.5281/zenodo.14580071}), OCT5k (\url{https://doi.org/10.5522/04/22128671.v4}), C8 (\url{https://www.kaggle.com/datasets/obulisainaren/retinal-oct-c8}), Kermany (\url{https://data.mendeley.com/datasets/rscbjbr9sj/2}), and OCTID (\url{https://borealisdata.ca/dataverse/OCTID}). The in-house training data used to fine-tune RetGrounder cannot be shared due to  data-sharing restrictions.

\section{Code availability}

A self-contained pipeline that reproduces the visual grounding evaluation end-to-end---dispatching inference for every evaluated open-source model family, parsing their predictions, and computing AggIoU from a single command---is available at \url{https://github.com/ronnief1/RetinaGroundBench}. All 27 evaluated open-source models are publicly available on Hugging Face; their exact repository identifiers are listed in Supplementary Table~\ref{tab:model_repos}.

\section{Competing interests}

The authors declare no competing interests.

\section{Acknowledgements}
This research was funded in part by the Austrian Science Fund (FWF) Grant-DOI:10.55776/FG9,
Christian Doppler Research Association, Austrian Federal Ministry of Economy, Energy and
Tourism, and the National Foundation for Research, Technology and Development.

\bibliographystyle{naturemag}
\bibliography{references}

\setcounter{figure}{0}
\setcounter{table}{0}

\renewcommand{\figurename}{Supplementary Fig.}

\renewcommand{\tablename}{Supplementary Table}

\newcounter{supptab}
\newcounter{suppfig}

\newcommand{\supptablabel}[1]{\refstepcounter{supptab}\def\@currentlabel{\thesupptab}\label{#1}}
\newcommand{\suppfiglabel}[1]{\refstepcounter{suppfig}\def\@currentlabel{\thesuppfig}\label{#1}}

\supptablabel{tab:model_repos}
\supptablabel{tab:perclass_aggiou}
\supptablabel{tab:perclass_f1}
\supptablabel{tab:perclass_precision}
\supptablabel{tab:perclass_recall}
\supptablabel{tab:perclass_aggiou_84}
\supptablabel{tab:perclass_f1_84}
\supptablabel{tab:prompts}
\supptablabel{tab:internal_grounding}
\supptablabel{tab:training_data}

\end{document}